\documentclass[11pt]{article}

\usepackage[preprint]{acl}

\usepackage{times}
\usepackage{latexsym}
\usepackage{amsmath}
\usepackage[T1]{fontenc}

\usepackage[utf8]{inputenc}

\usepackage{microtype}

\usepackage{inconsolata}

\usepackage{graphicx}
\usepackage{array}      
\usepackage{tabularx}
\usepackage{amsfonts}
\usepackage{booktabs}
\usepackage{tabularx}
\usepackage{multirow} 
\usepackage{subcaption}
\title{Distributional Semantics Tracing: A Framework for Explaining Hallucination in Large Language Models}

\author{Gagan Bhatia\textsuperscript{1}\,
Somayajulu G Sripada\textsuperscript{1} \,
Kevin Allan\textsuperscript{1}\,
Jacobo Azcona\textsuperscript{1}\,\\[0.4em]
  \textsuperscript{1}University of Aberdeen \,\\
  \texttt{\{g.bhatia.24,yaji.sripada\}@abdn.ac.uk}
  }

\begin{document}
\maketitle
\begin{abstract}

Hallucinations in large language models (LLMs) produce fluent continuations that are not supported by the prompt, especially under minimal contextual cues and ambiguity. We introduce \textbf{Distributional Semantics Tracing (DST)}, a model-native method that builds \emph{layer-wise semantic maps} at the answer position by decoding residual-stream states through the unembedding, selecting a compact top-$K$ concept set, and estimating directed concept-to-concept support via lightweight causal tracing. Using these traces, we test a representation-level hypothesis: hallucinations arise from \textbf{correlation-driven representational drift} across depth, where the residual stream is pulled toward a locally coherent but context-inconsistent concept neighborhood reinforced by training co-occurrences. On Racing Thoughts dataset, DST yields more faithful explanations than attribution, probing, and intervention baselines under an LLM-judge protocol, and the resulting \textbf{Contextual Alignment Score (CAS)} strongly predicts failures, supporting this drift hypothesis.

\end{abstract}

\section{Introduction}

Large language models (LLMs) can produce fluent outputs that are not supported by the prompt, including factual hallucinations, incorrect disambiguations under minimal contextual cues, and failures to follow counterfactual premises \citep{ji2023survey,dziri2022origin,tu2020empirical}. These behaviors persist in both standard generation and retrieval-augmented settings, where models may still contradict provided evidence or over-rely on parametric associations \citep{sun2024redeep,yu2024mechanistic}. Recent work also suggests that internal states contain measurable signals of hallucination risk and uncertainty before generation, indicating that failures are often detectable in the forward pass \citep{ji2024llm,orgad2024llms,wang2025what,zhang2025reasoning}. For practical debugging and scientific understanding, three questions are central: \emph{when} an incorrect continuation becomes detectable across layers, and \emph{what} human-comprehensible semantic representations can be computed from the model's latent distributional semantics and \emph{how} these can be used to construct a layer-wise semantic trace of the model internal mechanics from the final error back to the input layer.
\begin{figure*}[t]
    \centering
    \includegraphics[width=\textwidth]{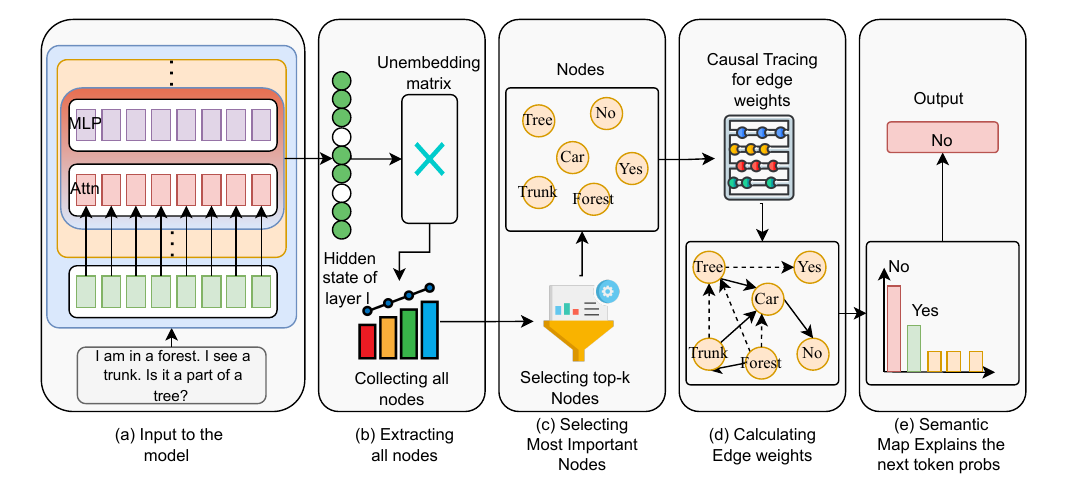}
    \caption{\textbf{DST pipeline overview.} (a) A prompt is processed by a decoder-only transformer. (b) At each layer, we decode the residual stream at the answer position through the unembedding to score vocabulary concepts. (c) We keep the top-$K$ concepts as nodes (detokenized to words for display). (d) We assign directed edge weights via causal tracing: for each upstream concept $v$, we minimally corrupt the prompt evidence most responsible for $v$ and measure the resulting change in probability of downstream concept $w$ at the answer position. The resulting semantic map summarizes which concept neighborhood is being assembled and how it supports competing continuations. In the example, a \emph{car}-centered neighborhood (``car trunk'' sense) gains support and links to the \emph{No} label despite the surrounding \emph{forest}/\emph{tree} context. Solid (dotted) edges denote positive (negative) support. (e) The model outputs the next token that is explained by the semantic map.}
    \label{fig:dst_pipeline}
\end{figure*}
Answering these questions requires interpretability tools that expose semantic content of intermediate representations with minimal assumptions. Token-level attribution methods, such as attention-based saliency \citep{vaswani2017attention}, LIME \citep{ribeiro2016should}, and Integrated Gradients \citep{sundararajan2017axiomatic}, can highlight prompt tokens correlated with a prediction, but they do not directly recover the model’s layer-wise latent \emph{distributional semantics}. Probing methods such as Logit Lens decode intermediate residual representations through the unembedding to inspect evolving next-token preferences \citep{wang2025logitlens4llmsextendinglogitlens}, but their outputs are typically lists of tokens rather than compact objects that summarize semantic structure and competing interpretations. Causal intervention approaches (e.g., activation patching and causal tracing) can localize influential components and layer ranges by measuring counterfactual logit effects \citep{meng2023locatingeditingfactualassociations,ameisen2025circuit}, and patch-based frameworks can elicit natural-language readouts of representations by controlled cross-context manipulations \citep{ghandeharioun2024patchscopes}. However, intervention-based methods often require multiple runs, carefully constructed clean/corrupted inputs, and additional experimental design choices that limit their use as lightweight diagnostics for a single generation.


This paper introduces \textbf{Distributional Semantics Tracing (DST) (Figure~\ref{fig:dst_pipeline})}, a model-native procedure that yields an interpretable \emph{layer-wise semantic map} for a single prompt by tracing how answer-position semantics evolve across layers. DST treats the residual stream at the answer position as a layer-indexed meaning representation and repeatedly performs two operations at each layer: (i) it projects the residual stream through the model’s unembedding to obtain a vocabulary-level compatibility signal (a logit-lens-style readout) \citep{wang2025logitlens4llmsextendinglogitlens}, and (ii) it summarizes the resulting neighborhood as a compact weighted graph whose nodes are the top-$K$ compatible concepts and whose directed edges quantify causal support between concepts via minimal prompt perturbations. Concretely, edges measure how removing the prompt evidence most responsible for concept $v$ reduces the probability assigned to concept $w$ at the answer position, yielding a lightweight causal trace over retrieved concepts.


In summary, this work makes three contributions:
\begin{itemize}
    \item We introduce \textbf{Distributional Semantics Tracing (DST)}, which produces \textbf{layer-wise semantic maps} as compact concept graphs derived from unembedding-based node retrieval and causal tracing for directed edge weights.
    \item We define \textbf{Contextual Alignment Score (CAS)} and operational layer markers that quantify representational drift and identify when semantic failures become detectable during the forward pass.
    \item We propose and empirically support a representation-level hypothesis for \textbf{why hallucinations occur}: failures arise from \textbf{correlation-driven representational drift}, where the residual stream is pulled toward a locally coherent but context-inconsistent concept neighborhood reinforced by training co-occurrences; DST semantic maps and CAS traces make this drift observable and predictive of error.
\end{itemize}

\section{Related Work}
\label{related_work}

Research into Large Language Model (LLM) fallibility has rapidly evolved from characterizing the problem of hallucination to developing a mechanistic understanding of its origins. 

\paragraph{Causes of Hallucination}
The challenge of hallucination, the generation of fluent yet factually inaccurate content, was identified early in the scaling of LLMs and is now recognised as a primary barrier to their reliable deployment \citep{ji2023survey, zhang2023siren, huang2023a, tonmoy2024a, bai2024hallucination, venkit2024confidently, cleti2024hallucinations}. Early research focused on characterizing and benchmarking this phenomenon from a black-box perspective, developing a taxonomy of errors \citep{zhang2023language, nan2021entity, hao2025beyond, walters2023fabrication} and creating evaluation suites to measure factual accuracy and consistency \citep{goodrich2019assessing, deyoung2020eraser, min2023factscore, li2023haluevallargescalehallucinationevaluation, ravichander2025halogen, zhang2024toolbehonest}. This work identified multiple root causes, including noise and biases in vast web-scale training corpora \citep{penedo2023refinedweb, penedo2024the, dolma, dziri2022origin}, a brittle and often superficial memorization of factual knowledge \citep{dankers2024generalisation, huang2024demystifying, stoehr2024localizing, haviv2023understanding, lu2024scaling, zhu2024beyond}, and the tendency for models to adopt shortcut learning strategies instead of robust reasoning \citep{geirhos2020shortcut, yuan2024llms, tang2023large, mccoy2019right, lai2021machine, niven2019probing}. In response, a diverse ecosystem of mitigation strategies has been developed. These include inference-time interventions such as grounding outputs with external data via Retrieval-Augmented Generation (RAG) \citep{lee2022factuality, ren2023investigating, huo2023retrieving, sun2024redeep, su2024mitigating, liang2024thames} and prompting models to perform self-correction and verification \citep{dhuliawala2023chain, zhang2024self, manakul2023selfcheckgpt, li2025two, sanwal2025layered, chu2025domaino1s, cheng2025thinkmorehallucinateless, lin2024interpreting}.
While effective at reducing symptoms, these methods largely operate on model inputs and outputs, leaving the internal mechanisms that produce hallucinations unaddressed \citep{hu2025fine-tuning, wei2023simple}.

\paragraph{Mechanisms of Interpretability}
To move beyond black-box corrections, our work leverages mechanistic interpretability (MI), a discipline focused on reverse-engineering the internal algorithms of neural networks \citep{olah2020zoom, bereska2024mechanistic, zhao2024towards, lin2025a, palikhe2025towards, singh2024rethinking}. This approach differs from classical XAI methods like LIME \citep{ribeiro2016""why, lime} or SHAP \citep{lundberg2017a, scott2017unified, amara2024syntaxshap} by causally analyzing model components. The MI toolkit, including causal tracing to find circuits \citep{meng2023locatingeditingfactualassociations, wang2022interpretabilitywildcircuitindirect, ameisen2025circuit, zhang2025mechanistic, harrasse2025tinysql, ou2025how, zhang2025finite}, dictionary learning with Sparse Autoencoders (SAEs) to uncover monosemantic features \citep{bricken2023towards, cunningham2023sparseautoencodershighlyinterpretable, minegishi2025rethinking}, and techniques like patching and the Logit Lens to inspect hidden states \citep{ghandeharioun2024patchscopes, wang2025logitlens4llmsextendinglogitlens, belrose2023elicitinglatentpredictionstransformers}, has revealed that LLMs learn coherent internal representations. Discoveries include how MLPs store factual knowledge \citep{geva2021transformer, chughtai2024summing}, how models perform multi-hop reasoning \citep{yang2024do}, and how they process multilingual inputs \citep{schut2025do, wendler2024do, saji2025romanlens}. This granular understanding is now being applied to diagnose failure modes like hallucination \citep{yu2024mechanistic, sun2024redeep, jiang2024devils}, offering a path to control model behavior directly through techniques like representation engineering \citep{zou2025representationengineeringtopdownapproach, bartoszcze2025representation, hu2025following, cywiński2025towards}. We posit that the failures MI uncovers are often consequences of architectural trade-offs made to achieve efficient scaling, a subject of extensive research covering everything from Mixture-of-Experts/Depths \citep{fedus2022switch, jiang2024mixtral, raposo2024mixture, elhoushi2024layerskip} and KV cache compression \citep{xiao2023efficient, ge2023model, zhang2023h2o, liu2023scissorhands} to looped, recurrent computation \citep{dehghani2018universal, giannou2023looped, saunshi2025reasoning}. 

\section{Tracing Semantic Failures}
\label{sec:tsf}

This section addresses three questions: (i) how to recover an interpretable, layer-wise view of a model’s distributional semantics for a single generation, (ii) when an incorrect continuation becomes detectable during a forward pass, and (iii) what representational dynamics precede hallucinations. Our core object is a \emph{layer-wise semantic map}: a compact weighted graph whose nodes are human-readable concepts most supported by the model’s residual stream at a given layer, and whose edges summarize coherence among those concepts under the model’s learned embedding geometry. We refer to the construction procedure as \textbf{Distributional Semantics Tracing (DST)}.

\subsection{Distributional Semantics Tracing (DST)}
\label{sec:dst}

A decoder-only transformer maintains a residual stream vector at every token position and layer. DST treats the residual stream at the \emph{answer position} as a moving ``meaning vector'' and repeatedly asks: \emph{(a) which vocabulary-level concepts are most compatible with the current representation \footnote{Throughout this section, we use \emph{representation} to refer specifically to the answer-position residual-stream state at layer $\ell$, $h^{\ell}_{i^\star}\in\mathbb{R}^d$, taken immediately prior to unembedding. We use \emph{unembedding geometry} to refer to the fixed vocabulary vectors $\{U_v\}$ (rows of the unembedding matrix $U$) and their induced similarity structure; this geometry is used only for linear readout and concept-to-concept comparisons, and is not the traced representation itself.}, and (b) which of those concepts form a contextually aligned neighborhood under the model’s embedding geometry?} DST answers these questions using only model-native linear operations (projection into the unembedding space, nearest-neighbor retrieval, and similarity computations), producing a compact semantic map per layer.
Let $x=(t_1,\dots,t_n)$ be the prompt tokens and let $f$ be a frozen decoder-only transformer with $L$ layers and residual width $d$. Let $h_i^\ell \in \mathbb{R}^d$ denote the residual stream at layer $\ell$ and token position $i$ (immediately before unembedding). Let $i^\star$ denote the \emph{answer position}, i.e., the position used to predict the next token. We analyze the sequence of residual states $\{h_{i^\star}^\ell\}_{\ell=1}^{L}$.

\noindent\paragraph{Step 1: project hidden states into a concept space.}
DST converts $h_{i^\star}^\ell$ into a vocabulary-level compatibility signal by scoring each vocabulary item via the model’s unembedding geometry. Let $U\in\mathbb{R}^{|\mathcal{V}|\times d}$ be the unembedding matrix (tied or untied). We define the concept score
\begin{equation}
\label{eq:concept_score}
s^\ell(v;i^\star) \;=\; \langle U_v,\; h_{i^\star}^\ell\rangle,
\end{equation}
which can be read as a compatibility between the current representation and the concept $v$. One alternative is to use SAEs \cite{cunningham2023sparseautoencodershighlyinterpretable} to decompose residual-stream features into a learned sparse dictionary; however, in our controlled minimal-pair setting, a direct Logit-Lens projection through the unembedding already yields a stable, human-readable top-K concept neighborhood at each layer, so we use this simpler model-native readout to recover layer-wise semantics without introducing an additional learned representation (and its training/hyperparameter dependencies).


\noindent\paragraph{Step 2: select a set of concept nodes.}
A full vocabulary distribution is not directly interpretable, so we form a node set by taking the top-$K$ concepts under $s^\ell(\cdot;i^\star)$:
\begin{equation}
\label{eq:topk_nodes}
V^\ell \;=\; \mathrm{TopK}\big(\{s^\ell(v;i^\star)\}_{v\in\mathcal{V}},\,K\big).
\end{equation}
Here, a \emph{node} is a human-readable concept displayed as a word: we merge adjacent subword vocabulary items that detokenize to the same surface word (aggregating their compatibility for display) so that nodes correspond to words rather than tokenizer fragments. We choose $K$ using the benchmark supervision available in \textbf{Racing Thoughts} (gold continuation, counterfactual foil continuation, and label tokens such as \emph{Yes}/\emph{No}) to ensure these target alternatives are captured within the retrieved neighborhood, and we keep $K$ small to preserve interpretability and control graph complexity, since the number of potential edges grows as $O(K^2)$ and large $K$ yields dense, noisy maps.

\noindent\paragraph{Step 3: compute edge strengths via causal tracing.}
Rather than measuring concept-to-concept coherence purely geometrically, we define directed edges \emph{causally} by testing whether the prompt evidence that most supports one concept is also \emph{necessary} for supporting another. For each node $v\in V^\ell$, we select an influential prompt position $p^\ell(v)$ (the prompt token position whose layer-$\ell$ residual state yields the highest unembedding score for $v$ under Eq.~\eqref{eq:concept_score}), construct a minimally corrupted prompt $\tilde x_{p^\ell(v)}$ by replacing the token at $p^\ell(v)$ with an unrelated token, and re-run the model to obtain the next-token distribution at the answer position. We then define the directed edge weight as the drop in probability assigned to $w$ under this corruption:
\begin{equation}
\label{eq:affinity_causal}
\Omega^\ell(v\Rightarrow w)
\;=\;
P\!\bigl(t_w \mid x\bigr)\;-\;P\!\bigl(t_w \mid \tilde x_{p^\ell(v)}\bigr),
\end{equation}
where $P(\cdot\mid x)=\mathrm{softmax}(z)$ is the next-token distribution from the original prompt logits $z$ at position $i^\star$, $P(\cdot\mid \tilde x_{p^\ell(v)})=\mathrm{softmax}(\tilde z)$ is the distribution after corruption (yielding logits $\tilde z$ at $i^\star$), and $t_w$ is the vocabulary index of concept $w$. Intuitively, $\Omega^\ell(v\Rightarrow w)$ is large when the evidence that most activates $v$ is also causally responsible for supporting $w$; chaining high-weight edges yields a multi-hop causal path that summarizes which prompt cues are stitched together before the model commits to the final continuation (e.g., \emph{Yes}/\emph{No}). Another alternative is full circuit tracing (head/MLP-level localization with many interventions), but we use minimal token-corruption causal effects to define edges because it provides the needed “which prompt evidence is necessary for which concept” signal while keeping the procedure lightweight and scalable.


\begin{figure*}[!ht]
    \centering
    \includegraphics[width=0.8\linewidth]{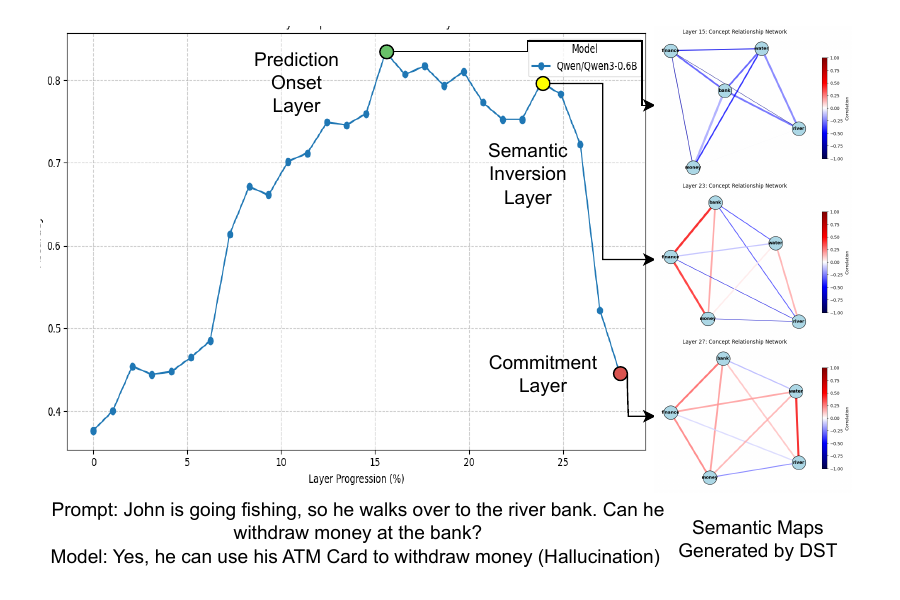}
    \caption{\textbf{Layer-wise onset of a semantic failure.} Left: contextual alignment score (CAS) across depth for Qwen/Qwen3-0.6B using a controlled ambiguity prompt (\emph{bank} disambiguated by \emph{river}), with markers for prediction onset (\textcolor{green!90!black}{Green} dot) , semantic inversion (\textcolor{yellow!90!black}{Yellow} dot) , and commitment (\textcolor{red!90!black}{Red} dot) . Right: DST semantic maps at representative layers show the corresponding structural shift from a river-aligned neighborhood to a finance-aligned neighborhood, explaining why probability mass ultimately concentrates on the incorrect financial continuation.}
    \label{fig:dst_layer_onset}
\end{figure*}

\noindent\paragraph{Step 4: relate the semantic map to next-token probabilities.}

DST is not a mechanistic decomposition of the full forward pass; rather, it provides an interpretable summary of (i) which concepts are locally compatible with the representation (Steps 1--2), and (ii) which concept-to-concept transitions are causally supported by specific prompt tokens (Step 3). We read the resulting graph alongside the model’s next-token distribution at the answer position: when the highest-probability continuation (e.g., the context-consistent \emph{Yes}/\emph{No} label or gold entity) lies at the terminus of a strong causal path whose intermediate nodes are themselves supported by cue-aligned evidence, the model typically produces the correct continuation. Conversely, failures occur when the strongest causal paths terminate in a competing neighborhood (e.g., the wrong sense or correlated entity), indicating that perturbing the key evidence for upstream nodes disproportionately reduces probability mass on the context-consistent alternative while leaving the competing alternative relatively intact. In this way, the causal edge weights provide a concrete link between prompt evidence, intermediate retrieved concepts, and the final next-token probabilities that determine the model’s output.

\subsection{Evaluating Explanation Faithfulness}
\label{sec:evaluating}

We compare DST to prior interpretability methods by translating each method’s internal signal into a short natural-language explanation and scoring explanations for \textbf{faithfulness} using an LLM-as-judge protocol. We focus on \textbf{Racing Thoughts} \citep{lepori2024racing}, which provides controlled minimal pairs where a small contextual cue determines the intended interpretation, making it well-suited for evaluating whether an explanation correctly identifies (i) the disambiguating cue and (ii) how the model did or did not use it.

\noindent\textbf{Interpretability methods evaluated.}
We evaluate DST against baselines spanning \emph{attribution}, \emph{probing}, and \emph{causal intervention} (Table~\ref{tab: results_rt}). Attribution baselines include attention saliency \citep{vaswani2017attention}, LIME \citep{ribeiro2016should}, and gradient-based path attribution (Integrated Gradients) \citep{sundararajan2017axiomatic}, as well as ReAGent \citep{zhao2024reagent}. For probing, we use Logit Lens \cite{wang2025logitlens4llmsextendinglogitlens}, decoding intermediate residual states through the unembedding to inspect layer-wise token preferences. Intervention baselines include Patchscopes \citep{ghandeharioun2024patchscopes} and activation patching / causal tracing \citep{meng2023locatingeditingfactualassociations,ameisen2025circuit}. We also compare to Subsequence Tracing, which identifies causal subsequences via randomized-context association tests \citep{sun2025why}. 


\noindent\textbf{Normalizing explanations across heterogeneous methods.}
Because methods return different primitives (token attributions, layer-wise logit predictions, activation trajectories, intervention sensitivities), we enforce a normalization: each explanation must cite concrete evidence at the granularity of the method’s output under the same length budget. Token-level methods cite the top-$k$ influential tokens/spans; Logit Lens-style probes cite the layer (or layer range) where preferences diverge; intervention methods cite the most sensitive layer range and prompt fragment; DST cites dominant nodes/edges in the semantic map and the earliest layer where contextual alignment begins to degrade (defined below).

\noindent\textbf{Judge protocol.}
For each example, the judge is shown: (i) the prompt, (ii) the model’s generated answer, (iii) the gold label, and (iv) one candidate explanation from a single method (randomized order across examples). Judges assign a 0--10 faithfulness score using a rubric emphasizing whether the explanation identifies the \emph{specific contextual cue} that should control the interpretation in Racing Thoughts, and whether it correctly characterizes how that cue influenced the model’s internal state and final prediction. We performed a small human evaluation which is explained in Appendix~\ref{sec:appendix_validation} and the prompt for evaluation and rubrics are presented in Appendix~\ref{app:judge_prompt}.

\subsection{Results: DST Produces More Faithful Traces}
\label{sec:results}

\begin{table}[!htp]
\centering
\scriptsize
\renewcommand{\arraystretch}{1.2} 
\setlength{\tabcolsep}{5pt} 
\resizebox{\columnwidth}{!}{%
\begin{tabular}{clccccc}
\toprule
\textbf{Type} & \textbf{Method} & \textbf{SmolLM2} & \textbf{Qwen3} & \textbf{OLMo2} & \textbf{Llama3.2} & \textbf{AVG} \\
 & & \textbf{135M} & \textbf{0.6B} & \textbf{1B} & \textbf{1B} & \\
\midrule

\multirow{4}{*}{\rotatebox{90}{Baseline}} 
 & Attention & 0.18 & 0.25 & 0.35 & 0.12 & 0.23 \\
 & LIME & 0.28 & 0.28 & 0.19 & 0.27 & 0.25 \\
 & Grad-SHAP & 0.33 & 0.37 & 0.31 & 0.28 & 0.32 \\
 & ReAGent & 0.38 & 0.46 & 0.29 & 0.33 & 0.37 \\
\midrule

\multirow{5}{*}{\rotatebox{90}{Advanced}} 
 & Logit Lens & 0.56 & 0.43 & 0.50 & 0.48 & 0.49 \\
 & Patchscopes & 0.58 & 0.51 & 0.46 & 0.51 & 0.52 \\
 & SAE & 0.58 & 0.64 & 0.43 & 0.52 & 0.54 \\
 & Subseq. Tracing & 0.55 & 0.55 & 0.59 & 0.55 & 0.56 \\
 & Causal Tracing & 0.60 & 0.57 & 0.58 & 0.59 & 0.59 \\
\midrule

\rotatebox{90}{Ours} & \textbf{DST} & \textbf{0.72} & \textbf{0.68} & \textbf{0.75} & \textbf{0.69} & \textbf{0.71} \\
\bottomrule
\end{tabular}}
\caption{Faithfulness scores on the Racing Thoughts benchmark across four compact language models. We compare our method, \textbf{Distributional Semantics Tracing (DST)}, against ten baselines: attention saliency \citep{vaswani2017attention}, LIME \citep{ribeiro2016should}, Gradient-SHAP (Integrated Gradients) \citep{sundararajan2017axiomatic}, ReAGent \citep{zhao2024reagent}, Token Evolution via Logit Lens \citep{wang2025logitlens4llmsextendinglogitlens}, Patchscopes \citep{ghandeharioun2024patchscopes}, Sparse Autoencoders (SAE) \cite{cunningham2023sparseautoencodershighlyinterpretable}, Subsequence Tracing \citep{sun2025why}, and Causal Path Tracing \citep{meng2023locatingeditingfactualassociations,ameisen2025circuit}. DST achieves the highest average faithfulness score (0.71).}
\label{tab: results_rt}
\end{table}
Table~\ref{tab: results_rt} reports mean faithfulness on Racing Thoughts across four compact models. (We also evaluated our setups on the Halogen \cite{ravichander2025halogen} dataset. Please see Appendix \ref{sec:halogen}.) Standard attribution methods score lowest, consistent with the observation that they often surface plausible-looking tokens (e.g., the ambiguous word itself) without capturing how the model resolves the ambiguity through context. More mechanistic approaches improve substantially, especially when they localize the layer range where representations become sensitive to the misleading interpretation. DST achieves the highest mean faithfulness across models in our setting, which we attribute to two properties: (i) it explicitly surfaces the \emph{concept neighborhood} that the representation occupies at each layer, and (ii) it provides a simple scalar trace of contextual alignment that pinpoints when semantics begin to drift. In qualitative analysis, DST explanations are also comparatively stable across prompts: they cite consistent evidence (dominant nodes/edges plus a drift onset layer) rather than switching across incompatible evidence types.

\subsection{When do hallucinations start?}
\label{sec:when_start}

Hallucinations are seldom introduced only at the final layer. Instead, we often observe a gradual shift across depth: early layers retrieve multiple plausible interpretations, while later layers increasingly favor one neighborhood. DST makes this visible by constructing a concept graph $G^\ell$ at each layer and tracking a scalar measure of whether the representation remains aligned with the context-consistent interpretation or drifts toward a competing one.

\noindent\textbf{Contextual Alignment Score (CAS).}
For each prompt in our evaluation benchmarks, we know the context-consistent continuation and the controlled cue(s) that specify the intended interpretation. At each layer, we partition retrieved concepts into two sets,
\[
V^\ell = V^\ell_{\text{ctx}} \cup V^\ell_{\text{nonctx}},
\]
where $V^\ell_{\text{ctx}}$ contains concepts compatible with the intended interpretation and $V^\ell_{\text{nonctx}}$ contains concepts aligned with the competing interpretation.
Let $e(v)=U_v/\|U_v\|$ denote the normalized unembedding direction for concept $v$, and define the (signed) concept alignment at layer $\ell$ as
\begin{equation}
\label{eq:affinity_a}
a^\ell(v) \;=\; \cos\big(h_{i^\star}^\ell,\; e(v)\big).
\end{equation} 
Here, $h_{i^\star}^\ell$ is the residual stream at layer $\ell$ and the \emph{answer position} $i^\star$ (i.e., the final prompt token position whose representation is used to predict the next token).
We define CAS as the fraction of total absolute alignment assigned to the context-consistent set:
\begin{equation}
\label{eq:cas}
\mathrm{CAS}^\ell \;=\;
\frac{\sum_{v\in V^\ell_{\text{ctx}}} |a^\ell(v)|}
{\sum_{v\in V^\ell_{\text{ctx}} \cup V^\ell_{\text{nonctx}}} |a^\ell(v)|}.
\end{equation}
High CAS indicates that the representation aligns more strongly with context-consistent concepts; low CAS indicates increasing alignment with a competing interpretation.

\noindent\textbf{Operational layer markers.}
CAS yields three lightweight, operational markers that mirror the qualitative evolution in the semantic maps (as visulaised in Figure~\ref{fig:dst_layer_onset}. We define the \emph{prediction onset layer} (\textcolor{green!90!black}{Green} dot) as the earliest layer where CAS begins a sustained decline relative to the immediately preceding layer (a drop exceeding a small fixed tolerance), indicating the first consistent shift away from the context-consistent neighborhood. We define the \emph{semantic inversion layer} (\textcolor{yellow!90!black}{Yellow} dot) as the first layer where CAS falls below a fixed threshold (we use $0.8$) indicating a substantial takeover by the competing neighborhood. Finally, we define the \emph{commitment layer} (\textcolor{red!90!black}{Red} dot) as the first layer after inversion beyond which CAS remains persistently low through the remaining depth (below a fixed threshold), indicating that the model has effectively locked into the competing interpretation rather than returning to the context-consistent neighbourhood. 


In Figure~\ref{fig:dst_layer_onset} the prompt is a controlled ambiguity (\emph{bank}) where the cue \emph{river} specifies the intended river-bank sense, but the model answers with the financial sense. The left panel plots $\mathrm{CAS}^\ell$ across depth and marks three phases that align with the semantic maps on the right: (i) at the \emph{prediction onset} layer (green), $\mathrm{CAS}^\ell$ peaks and begins a sustained decline, indicating the first consistent shift away from river-aligned concepts; (ii) at the \emph{semantic inversion} layer (yellow), $\mathrm{CAS}^\ell$ crosses the threshold as the competing semantic interpretation  becomes dominant; and (iii) at the \emph{commitment} layer (red), $\mathrm{CAS}^\ell$ collapses and remains low, reflecting lock-in. Consistent with this trace, the Layer 15 map remains organized around river-context nodes (e.g., \emph{river}, \emph{water}) connected through \emph{bank}, whereas by Layers 23 and 27 the map increasingly concentrates around financial nodes (e.g., \emph{money}, \emph{finance}), with river-context nodes becoming weak or peripheral; this structural takeover explains why next-token probability ultimately concentrates on the incorrect financial continuation.

\begin{figure}[t]
    \centering
    \includegraphics[width=\columnwidth]{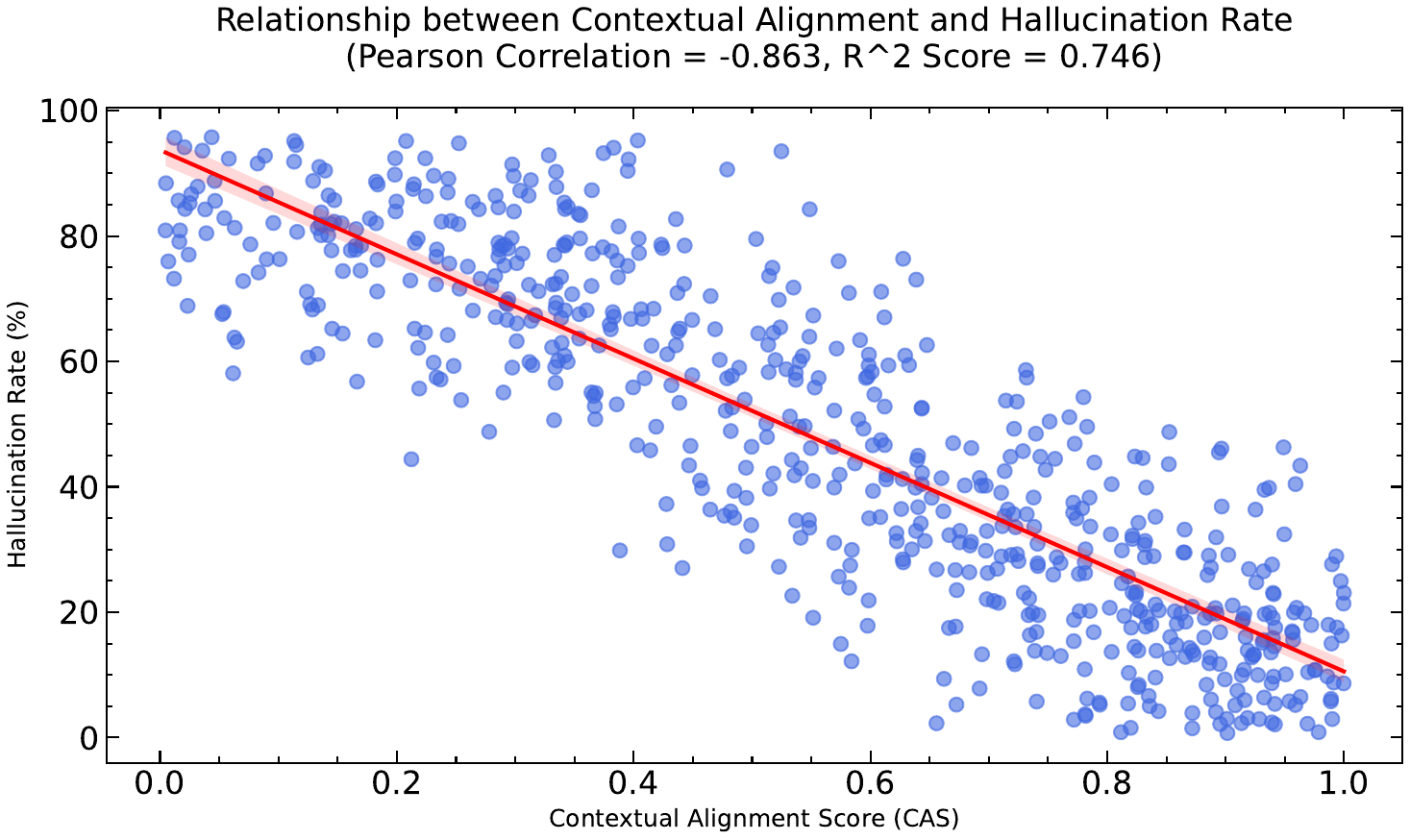}
    \caption{\textbf{CAS predicts hallucination rate.} Each point is an evaluation example with its final-layer CAS and whether the model hallucinated. Higher CAS corresponds to lower hallucination rate (Pearson $r=-0.863$; linear fit $R^2=0.746$).}
    \label{fig:cas_vs_hallucination}
\end{figure}

\begin{figure*}[!ht]
    \centering
    \includegraphics[width=0.8\linewidth]{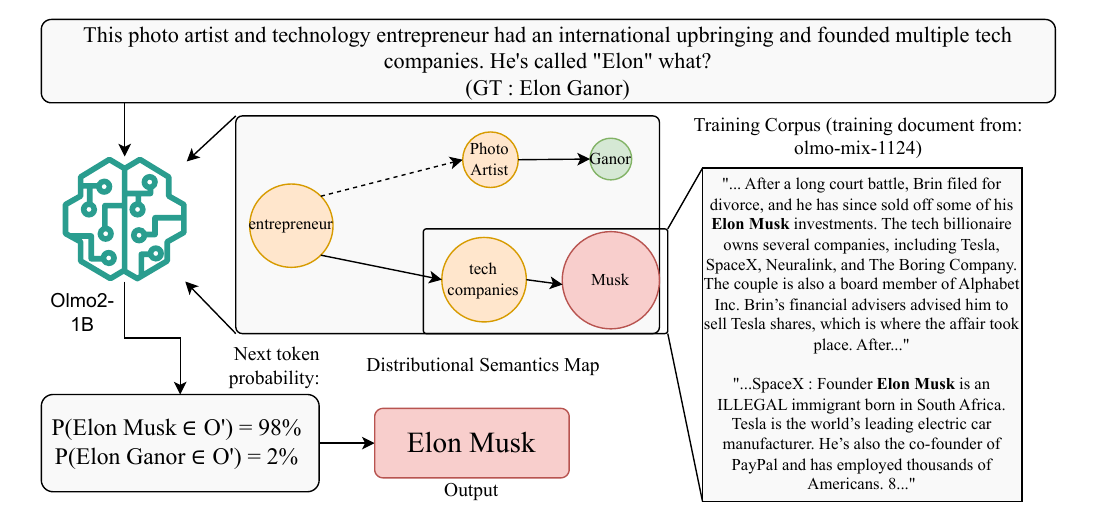}
    \caption{\textbf{Entity hallucination via correlation-driven drift.} The prompt describes the low-frequency entity \textit{Elon Ganor}, but the final-layer semantic map is dominated by a high-frequency correlated entity neighborhood around \textit{Musk}. The next-token distribution reflects this takeover (e.g., $P(\textit{Elon Musk}) \gg P(\textit{Elon Ganor})$): generic attributes in the prompt activate a concept cluster that is strongly reinforced by training co-occurrences, steering the residual stream toward the wrong entity despite the context specifying the correct one.}
    \label{fig:dst_entity_hijack}
\end{figure*}

\subsection{Why do hallucinations occur?}
\label{sec:why_occur}

\noindent\textbf{Correlation-driven representational drift.}
DST supports a single, representation-level account that ties together the method, the layer-wise onset analysis, and the qualitative failures: \textbf{hallucinations arise when the model’s internal representation is driven by spurious correlations between concepts rather than by the prompt’s contextual evidence}. Because next-token training rewards predictive co-occurrences, pretrained LMs can internalise correlations that are strong in the training distribution but not warranted in the current input, especially when the prompt contains generic descriptors, ambiguous cues, or counterfactual statements. Prior work on hallucinations and model brittleness emphasises the role of training-data artefacts and distributional mismatch in producing fluent but incorrect generations \citep{tu2020empirical,dziri2022origin,ji2023survey}.
In DST terms, this failure mode appears as a layer-wise shift in which concepts are most compatible with the residual stream at the answer position: early layers can reflect multiple plausible neighbourhoods, but later layers increasingly favour the correlated (and often context-inconsistent) neighbourhood, and the output distribution follows that choice. Our Contextual Alignment Score (CAS) operationalises this process: when the model begins to privilege the non-contextual neighborhood, CAS declines (onset), may cross the inversion threshold, and often stabilizes once the model has effectively committed to the correlated interpretation (commitment). Throughout this section, we report \textbf{hallucination rate} as the \emph{error rate} ($1-\text{accuracy}$) computed over all 750 evaluation examples per model from Table~\ref{tab: results_rt}.
Empirically, CAS is strongly predictive of failure rate across examples: Figure~\ref{fig:cas_vs_hallucination} shows a tight negative relationship between final-layer CAS and hallucination rate (Pearson $r=-0.863$, $R^2=0.746$), consistent with the claim that loss of contextual alignment is a primary driver of hallucinated outputs.

\noindent\textbf{Entity-setting failures as a concrete mechanism.}
Figure~\ref{fig:dst_entity_hijack} illustrates the mechanism concretely in an entity-setting failure. The prompt describes a relatively infrequent correct entity (GT: \textit{Elon Ganor}), but the final-layer distributional semantics map is dominated by a neighborhood containing generic attributes (e.g., \emph{photo artist}, \emph{technology entrepreneur}, \emph{founded multiple tech companies}) and a highly frequent correlated entity token (\textit{Musk}). The key point is not that the model ``ignores'' the prompt; rather, the prompt’s attributes activate internal associations that are strongly reinforced by training co-occurrence patterns (and, in practice, by noisy or misleading training passages that repeatedly pair those attributes with a famous entity), pulling the residual stream toward the wrong neighborhood \citep{dziri2022origin,ji2023survey}. Mechanistically, such correlations must be instantiated as internal feature-to-logit pathways that amplify particular associations and steer logits toward the correlated completion; causal intervention work shows that editing or patching internal activations can directly modify these associations and change the resulting prediction \citep{meng2023locatingeditingfactualassociations,ameisen2025circuit}. DST provides a compact way to \emph{observe} the consequence of these pathways during a single forward pass: as the correlated neighborhood becomes more prominent across layers, the next-token probability concentrates on the correlated entity (e.g., $P(\textit{Elon Musk}) \gg P(\textit{Elon Ganor})$ in Figure~\ref{fig:dst_entity_hijack}), yielding a fluent but incorrect answer. Taken together, these results support the paper’s central claim: hallucinations are not arbitrary decoding failures; they are the downstream result of correlation-driven representational drift that DST visualizes, and that CAS detects and predicts.

\section{Conclusion}
We introduced \emph{Distributional Semantics Tracing (DST)}, a model-native method that produces \emph{layer-wise semantic maps} for a single prompt by combining unembedding-based concept retrieval with lightweight causal tracing. We also defined the \emph{Contextual Alignment Score (CAS)} and operational layer markers (onset, inversion, commitment) that indicate \emph{when} semantic failures become detectable during the forward pass.
Future work will extend DST beyond hallucination to test whether similar layer-wise semantic dynamics underlie stereotypes and model bias \citep{10.1145/3582269.3615599} as well as broader forms of model misalignment \citep{QU2025249}, further evaluating DST as a general account of semantic failure in language models.

\section*{Limitations}
DST’s semantic maps depend on discrete top-$K$ unembedding retrieval and detokenization heuristics, which can miss relevant concepts outside the retrieved set, blur compositional or multi-token semantics, and become noisy as $K$ increases; moreover, its directed edges rely on a specific minimal-corruption operator and a procedure for selecting influential prompt positions, so edge magnitudes (and occasionally directions) may vary with alternative corruption schemes, distributed evidence across spans, or prompts where local token replacement is unnatural. Although DST is lighter than full clean/corrupted circuit analyses, Step~3 still incurs additional forward passes (one per retrieved concept, potentially across layers), which may limit scalability to large models or long contexts without batching or approximation. Our evaluation emphasizes controlled ambiguity minimal pairs and an LLM-as-judge faithfulness protocol, which, despite rubric validation, may not fully capture long-form factual hallucinations, tool-augmented settings, or human interpretive preferences, and it remains an open empirical question how onset/inversion/commitment dynamics generalize across substantially different architectures (e.g., MoE, recurrent/looped transformers) and retrieval pipelines that alter evidence flow.

\section*{Ethical Considerations}
DST is intended to improve reliability and transparency, but mechanistic diagnostics are inherently dual-use: the same causal sensitivities that help auditors and engineers localize drift could be leveraged to design more effective steering or exploitation strategies, so responsible dissemination should include clear use guidelines and safeguards for large-scale intervention sweeps. Because semantic maps are compelling artifacts, they can create an interpretability illusion: DST surfaces model-supported concept neighborhoods and counterfactual effects under a particular intervention design, not ground-truth reasoning, so practitioners should treat outputs as diagnostic evidence that requires corroboration rather than definitive explanations. DST may also reveal or operationalize biased associations embedded in training data, and traces computed on sensitive prompts could expose private or proprietary content through stored intermediate readouts, implying that deployments should pair DST with fairness audits, careful access controls, and data-handling practices (e.g., redaction and limited retention) appropriate for the sensitivity of analyzed inputs.

\section*{Broader Impact}
By providing compact layer-wise semantic objects and a scalar contextual-alignment trace, DST can shorten debugging cycles, support more informative evaluation than aggregate accuracy, and enable earlier, more targeted mitigations for hallucination by identifying the onset of representational drift before final commitment; scientifically, it offers a bridge between black-box hallucination taxonomies and mechanistic accounts by making depth-wise takeovers observable and testable, potentially informing architectural and training-objective choices that better preserve contextual grounding. In safety and governance contexts, DST-style evidence can improve failure documentation and auditing by linking prompt cues to internal drift trajectories, but it also risks misuse (e.g., adversarial steering) and misinterpretation (over-trust in visually coherent maps), underscoring the need for responsible use, conservative communication of scope and uncertainty, and evaluation across diverse tasks and populations to ensure that improved interpretability translates into broadly reliable and equitable behavior.

\bibliography{custom}

@article{zhang2025reasoning,
Author        = {Anqi Zhang and Yulin Chen and Jane Pan and Chen Zhao and Aurojit Panda and Jinyang Li and He He},
Title         = {Reasoning Models Know When They're Right: Probing Hidden States for
  Self-Verification},
journal        = {2504.05419v1},
ArchivePrefix = {arXiv},
PrimaryClass  = {cs.AI},
Year          = {2025},
Month         = {Apr},
Url           = {http://arxiv.org/abs/2504.05419v1}
}

@article{minegishi2025rethinking,
Author        = {Gouki Minegishi and Hiroki Furuta and Yusuke Iwasawa and Yutaka Matsuo},
Title         = {Rethinking Evaluation of Sparse Autoencoders through the Representation
  of Polysemous Words},
journal        = {2501.06254v2},
ArchivePrefix = {arXiv},
PrimaryClass  = {cs.CL},
Year          = {2025},
Month         = {Jan},
Url           = {http://arxiv.org/abs/2501.06254v2}
}

@article{schut2025do,
Author        = {Lisa Schut and Yarin Gal and Sebastian Farquhar},
Title         = {Do Multilingual LLMs Think In English?},
journal        = {2502.15603v1},
ArchivePrefix = {arXiv},
PrimaryClass  = {cs.CL},
Year          = {2025},
Month         = {Feb},
Url           = {http://arxiv.org/abs/2502.15603v1}
}

@article{jiang2024devils,
Author        = {Zhangqi Jiang and Junkai Chen and Beier Zhu and Tingjin Luo and Yankun Shen and Xu Yang},
Title         = {Devils in Middle Layers of Large Vision-Language Models: Interpreting,
  Detecting and Mitigating Object Hallucinations via Attention Lens},
journal        = {2411.16724v3},
ArchivePrefix = {arXiv},
PrimaryClass  = {cs.CV},
Year          = {2024},
Month         = {Nov},
Url           = {http://arxiv.org/abs/2411.16724v3}
}

@article{wendler2024do,
Author        = {Chris Wendler and Veniamin Veselovsky and Giovanni Monea and Robert West},
Title         = {Do Llamas Work in English? On the Latent Language of Multilingual
  Transformers},
journal        = {2402.10588v4},
ArchivePrefix = {arXiv},
PrimaryClass  = {cs.CL},
Year          = {2024},
Month         = {Feb},
Url           = {http://arxiv.org/abs/2402.10588v4}
}

@article{hu2025following,
Author        = {Ling Hu and Yuemei Xu and Xiaoyang Gu and Letao Han},
Title         = {Following the Whispers of Values: Unraveling Neural Mechanisms Behind
  Value-Oriented Behaviors in LLMs},
journal        = {2504.04994v2},
ArchivePrefix = {arXiv},
PrimaryClass  = {cs.CL},
Year          = {2025},
Month         = {Apr},
Url           = {http://arxiv.org/abs/2504.04994v2}
}

@article{ameisen2025circuit,
  author={Ameisen, Emmanuel and Lindsey, Jack and Pearce, Adam and Gurnee, Wes and Turner, Nicholas L. and Chen, Brian and Citro, Craig and Abrahams, David and Carter, Shan and Hosmer, Basil and Marcus, Jonathan and Sklar, Michael and Templeton, Adly and Bricken, Trenton and McDougall, Callum and Cunningham, Hoagy and Henighan, Thomas and Jermyn, Adam and Jones, Andy and Persic, Andrew and Qi, Zhenyi and Ben Thompson, T. and Zimmerman, Sam and Rivoire, Kelley and Conerly, Thomas and Olah, Chris and Batson, Joshua},
  title={Circuit Tracing: Revealing Computational Graphs in Language Models},
  journal={Transformer Circuits Thread},
  year={2025},
  url={https://transformer-circuits.pub/2025/attribution-graphs/methods.html}
}

@article{ghandeharioun2024patchscopes,
Author        = {Asma Ghandeharioun and Avi Caciularu and Adam Pearce and Lucas Dixon and Mor Geva},
Title         = {Patchscopes: A Unifying Framework for Inspecting Hidden Representations
  of Language Models},
journal        = {2401.06102v4},
ArchivePrefix = {arXiv},
PrimaryClass  = {cs.CL},
Year          = {2024},
Month         = {Jan},
Url           = {http://arxiv.org/abs/2401.06102v4}
}

@article{zhang2025finite,
Author        = {Yifan Zhang and Wenyu Du and Dongming Jin and Jie Fu and Zhi Jin},
Title         = {Finite State Automata Inside Transformers with Chain-of-Thought: A
  Mechanistic Study on State Tracking},
journal        = {2502.20129v3},
ArchivePrefix = {arXiv},
PrimaryClass  = {cs.CL},
Year          = {2025},
Month         = {Feb},
Url           = {http://arxiv.org/abs/2502.20129v3}
}

@article{zhao2024towards,
Author        = {Haiyan Zhao and Fan Yang and Bo Shen and Himabindu Lakkaraju and Mengnan Du},
Title         = {Towards Uncovering How Large Language Model Works: An Explainability
  Perspective},
journal        = {2402.10688v2},
ArchivePrefix = {arXiv},
PrimaryClass  = {cs.CL},
Year          = {2024},
Month         = {Feb},
Url           = {http://arxiv.org/abs/2402.10688v2}
}

@article{sanwal2025layered,
Author        = {Manish Sanwal},
Title         = {Layered Chain-of-Thought Prompting for Multi-Agent LLM Systems: A
  Comprehensive Approach to Explainable Large Language Models},
journal        = {2501.18645v2},
ArchivePrefix = {arXiv},
PrimaryClass  = {cs.CL},
Year          = {2025},
Month         = {Jan},
Url           = {http://arxiv.org/abs/2501.18645v2}
}

@article{chu2025domaino1s,
Author        = {Xu Chu and Zhijie Tan and Hanlin Xue and Guanyu Wang and Tong Mo and Weiping Li},
Title         = {Domaino1s: Guiding LLM Reasoning for Explainable Answers in High-Stakes
  Domains},
journal        = {2501.14431v2},
ArchivePrefix = {arXiv},
PrimaryClass  = {cs.CL},
Year          = {2025},
Month         = {Jan},
Url           = {http://arxiv.org/abs/2501.14431v2}
}

@article{li2025two,
Author        = {Jiazheng Li and Yuxiang Zhou and Junru Lu and Gladys Tyen and Lin Gui and Cesare Aloisi and Yulan He},
Title         = {Two Heads Are Better Than One: Dual-Model Verbal Reflection at
  Inference-Time},
journal        = {2502.19230v1},
ArchivePrefix = {arXiv},
PrimaryClass  = {cs.CL},
Year          = {2025},
Month         = {Feb},
Url           = {http://arxiv.org/abs/2502.19230v1}
}

@article{lin2025a,
Author        = {Zihao Lin and Samyadeep Basu and Mohammad Beigi and Varun Manjunatha and Ryan A. Rossi and Zichao Wang and Yufan Zhou and Sriram Balasubramanian and Arman Zarei and Keivan Rezaei and Ying Shen and Barry Menglong Yao and Zhiyang Xu and Qin Liu and Yuxiang Zhang and Yan Sun and Shilong Liu and Li Shen and Hongxuan Li and Soheil Feizi and Lifu Huang},
Title         = {A Survey on Mechanistic Interpretability for Multi-Modal Foundation
  Models},
journal        = {2502.17516v1},
ArchivePrefix = {arXiv},
PrimaryClass  = {cs.LG},
Year          = {2025},
Month         = {Feb},
Url           = {http://arxiv.org/abs/2502.17516v1}
}

@article{zhang2025mechanistic,
Author        = {Lin Zhang and Lijie Hu and Di Wang},
Title         = {Mechanistic Unveiling of Transformer Circuits: Self-Influence as a Key
  to Model Reasoning},
journal        = {2502.09022v2},
ArchivePrefix = {arXiv},
PrimaryClass  = {cs.AI},
Year          = {2025},
Month         = {Feb},
Url           = {http://arxiv.org/abs/2502.09022v2}
}

@article{ou2025how,
Author        = {Yixin Ou and Yunzhi Yao and Ningyu Zhang and Hui Jin and Jiacheng Sun and Shumin Deng and Zhenguo Li and Huajun Chen},
Title         = {How Do LLMs Acquire New Knowledge? A Knowledge Circuits Perspective on
  Continual Pre-Training},
journal        = {2502.11196v2},
ArchivePrefix = {arXiv},
PrimaryClass  = {cs.LG},
Year          = {2025},
Month         = {Feb},
Url           = {http://arxiv.org/abs/2502.11196v2}
}

@article{harrasse2025tinysql,
Author        = {Abir Harrasse and Philip Quirke and Clement Neo and Dhruv Nathawani and Luke Marks and Amir Abdullah},
Title         = {TinySQL: A Progressive Text-to-SQL Dataset for Mechanistic
  Interpretability Research},
journal        = {2503.12730v3},
ArchivePrefix = {arXiv},
PrimaryClass  = {cs.LG},
Year          = {2025},
Month         = {Mar},
Url           = {http://arxiv.org/abs/2503.12730v3}
}

@article{bartoszcze2025representation,
Author        = {Lukasz Bartoszcze and Sarthak Munshi and Bryan Sukidi and Jennifer Yen and Zejia Yang and David Williams-King and Linh Le and Kosi Asuzu and Carsten Maple},
Title         = {Representation Engineering for Large-Language Models: Survey and
  Research Challenges},
journal        = {2502.17601v1},
ArchivePrefix = {arXiv},
PrimaryClass  = {cs.AI},
Year          = {2025},
Month         = {Feb},
Url           = {http://arxiv.org/abs/2502.17601v1}
}

@article{chughtai2024summing,
Author        = {Bilal Chughtai and Alan Cooney and Neel Nanda},
Title         = {Summing Up the Facts: Additive Mechanisms Behind Factual Recall in LLMs},
journal        = {2402.07321v1},
ArchivePrefix = {arXiv},
PrimaryClass  = {cs.LG},
Year          = {2024},
Month         = {Feb},
Url           = {http://arxiv.org/abs/2402.07321v1}
}

@article{saji2025romanlens,
Author        = {Alan Saji and Jaavid Aktar Husain and Thanmay Jayakumar and Raj Dabre and Anoop Kunchukuttan and Ratish Puduppully},
Title         = {RomanLens: The Role Of Latent Romanization In Multilinguality In LLMs},
journal        = {2502.07424v3},
ArchivePrefix = {arXiv},
PrimaryClass  = {cs.CL},
Year          = {2025},
Month         = {Feb},
Url           = {http://arxiv.org/abs/2502.07424v3}
}

@article{bereska2024mechanistic,
Author        = {Leonard Bereska and Efstratios Gavves},
Title         = {Mechanistic Interpretability for AI Safety -- A Review},
journal        = {2404.14082v3},
ArchivePrefix = {arXiv},
PrimaryClass  = {cs.AI},
Year          = {2024},
Month         = {Apr},
Url           = {http://arxiv.org/abs/2404.14082v3}
}

@article{wang2025what,
Author        = {Peiran Wang and Yang Liu and Yunfei Lu and Jue Hong and Ye Wu},
Title         = {What are Models Thinking about? Understanding Large Language Model
  Hallucinations ""Psychology"" through Model Inner State Analysis},
journal        = {2502.13490v1},
ArchivePrefix = {arXiv},
PrimaryClass  = {cs.CL},
Year          = {2025},
Month         = {Feb},
Url           = {http://arxiv.org/abs/2502.13490v1}
}

@article{yu2024mechanistic,
Author        = {Lei Yu and Meng Cao and Jackie Chi Kit Cheung and Yue Dong},
Title         = {Mechanistic Understanding and Mitigation of Language Model Non-Factual
  Hallucinations},
journal        = {2403.18167v2},
ArchivePrefix = {arXiv},
PrimaryClass  = {cs.CL},
Year          = {2024},
Month         = {Mar},
Url           = {http://arxiv.org/abs/2403.18167v2}
}

@article{sun2024redeep,
Author        = {Zhongxiang Sun and Xiaoxue Zang and Kai Zheng and Yang Song and Jun Xu and Xiao Zhang and Weijie Yu and Yang Song and Han Li},
Title         = {ReDeEP: Detecting Hallucination in Retrieval-Augmented Generation via
  Mechanistic Interpretability},
journal        = {2410.11414v2},
ArchivePrefix = {arXiv},
PrimaryClass  = {cs.CL},
Year          = {2024},
Month         = {Oct},
Url           = {http://arxiv.org/abs/2410.11414v2}
}

@article{orgad2024llms,
Author        = {Hadas Orgad and Michael Toker and Zorik Gekhman and Roi Reichart and Idan Szpektor and Hadas Kotek and Yonatan Belinkov},
Title         = {LLMs Know More Than They Show: On the Intrinsic Representation of LLM
  Hallucinations},
journal        = {2410.02707v4},
ArchivePrefix = {arXiv},
PrimaryClass  = {cs.CL},
Year          = {2024},
Month         = {Oct},
Url           = {http://arxiv.org/abs/2410.02707v4}
}

@article{ji2024llm,
Author        = {Ziwei Ji and Delong Chen and Etsuko Ishii and Samuel Cahyawijaya and Yejin Bang and Bryan Wilie and Pascale Fung},
Title         = {LLM Internal States Reveal Hallucination Risk Faced With a Query},
journal        = {2407.03282v2},
ArchivePrefix = {arXiv},
PrimaryClass  = {cs.CL},
Year          = {2024},
Month         = {Jul},
Url           = {http://arxiv.org/abs/2407.03282v2}
}

@article{zhang2024toolbehonest,
Author        = {Yuxiang Zhang and Jing Chen and Junjie Wang and Yaxin Liu and Cheng Yang and Chufan Shi and Xinyu Zhu and Zihao Lin and Hanwen Wan and Yujiu Yang and Tetsuya Sakai and Tian Feng and Hayato Yamana},
Title         = {ToolBeHonest: A Multi-level Hallucination Diagnostic Benchmark for
  Tool-Augmented Large Language Models},
journal        = {2406.20015v2},
ArchivePrefix = {arXiv},
PrimaryClass  = {cs.CL},
Year          = {2024},
Month         = {Jun},
Url           = {http://arxiv.org/abs/2406.20015v2}
}

@article{lin2024interpreting,
Author        = {Zheng Lin and Zhenxing Niu and Zhibin Wang and Yinghui Xu},
Title         = {Interpreting and Mitigating Hallucination in MLLMs through Multi-agent
  Debate},
journal        = {2407.20505v1},
ArchivePrefix = {arXiv},
PrimaryClass  = {cs.CV},
Year          = {2024},
Month         = {Jul},
Url           = {http://arxiv.org/abs/2407.20505v1}
}

@article{sun2025why,
Author        = {Yiyou Sun and Yu Gai and Lijie Chen and Abhilasha Ravichander and Yejin Choi and Dawn Song},
Title         = {Why and How LLMs Hallucinate: Connecting the Dots with Subsequence
  Associations},
journal        = {2504.12691v1},
ArchivePrefix = {arXiv},
PrimaryClass  = {cs.CL},
Year          = {2025},
Month         = {Apr},
Url           = {http://arxiv.org/abs/2504.12691v1}
}

@article{ravichander2025halogen,
Author        = {Abhilasha Ravichander and Shrusti Ghela and David Wadden and Yejin Choi},
Title         = {HALoGEN: Fantastic LLM Hallucinations and Where to Find Them},
journal        = {2501.08292v1},
ArchivePrefix = {arXiv},
PrimaryClass  = {cs.CL},
Year          = {2025},
Month         = {Jan},
Url           = {http://arxiv.org/abs/2501.08292v1}
}

@article{lepori2024racing,
Author        = {Michael A. Lepori and Michael C. Mozer and Asma Ghandeharioun},
Title         = {Racing Thoughts: Explaining Contextualization Errors in Large Language
  Models},
journal        = {2410.02102v2},
ArchivePrefix = {arXiv},
PrimaryClass  = {cs.CL},
Year          = {2024},
Month         = {Oct},
Url           = {http://arxiv.org/abs/2410.02102v2}
}

@article{cywiński2025towards,
Author        = {Bartosz Cywiński and Emil Ryd and Senthooran Rajamanoharan and Neel Nanda},
Title         = {Towards eliciting latent knowledge from LLMs with mechanistic
  interpretability},
journal        = {2505.14352v1},
ArchivePrefix = {arXiv},
PrimaryClass  = {cs.LG},
Year          = {2025},
Month         = {May},
Url           = {http://arxiv.org/abs/2505.14352v1}
}

@article{yang2024do,
Author        = {Sohee Yang and Elena Gribovskaya and Nora Kassner and Mor Geva and Sebastian Riedel},
Title         = {Do Large Language Models Latently Perform Multi-Hop Reasoning?},
journal        = {2402.16837v2},
ArchivePrefix = {arXiv},
PrimaryClass  = {cs.CL},
Year          = {2024},
Month         = {Feb},
Url           = {http://arxiv.org/abs/2402.16837v2}
}

@article{ji2023survey,
	title        = {Survey of hallucination in natural language generation},
	author       = {Ji, Ziwei and Lee, Nayeon and Frieske, Rita and Yu, Tiezheng and Su, Dan and Xu, Yan and Ishii, Etsuko and Bang, Ye Jin and Madotto, Andrea and Fung, Pascale},
	year         = 2023,
	journal      = {ACM Computing Surveys},
	publisher    = {ACM New York, NY},
	volume       = 55,
	number       = 12,
	pages        = {1--38}
}

@article{zhang2023siren,
	title        = {Siren's song in the AI ocean: a survey on hallucination in large language models},
	author       = {Zhang, Yue and Li, Yafu and Cui, Leyang and Cai, Deng and Liu, Lemao and Fu, Tingchen and Huang, Xinting and Zhao, Enbo and Zhang, Yu and Chen, Yulong and others},
	year         = 2023,
	journal      = {arXiv prjournal arXiv:2309.01219}
}

@article{zhang2023language,
	title        = {How language model hallucinations can snowball},
	author       = {Zhang, Muru and Press, Ofir and Merrill, William and Liu, Alisa and Smith, Noah A},
	year         = 2023,
	journal      = {arXiv prjournal arXiv:2305.13534}
}

@article{dziri2022origin,
	title        = {On the origin of hallucinations in conversational models: Is it the datasets or the models?},
	author       = {Dziri, Nouha and Milton, Sivan and Yu, Mo and Zaiane, Osmar and Reddy, Siva},
	year         = 2022,
	journal      = {arXiv prjournal arXiv:2204.07931}
}

@article{penedo2023refinedweb,
	title        = {The RefinedWeb dataset for Falcon LLM: outperforming curated corpora with web data, and web data only},
	author       = {Penedo, Guilherme and Malartic, Quentin and Hesslow, Daniel and Cojocaru, Ruxandra and Cappelli, Alessandro and Alobeidli, Hamza and Pannier, Baptiste and Almazrouei, Ebtesam and Launay, Julien},
	year         = 2023,
	journal      = {arXiv prjournal arXiv:2306.01116}
}

@article{ren2023investigating,
	title        = {Investigating the factual knowledge boundary of large language models with retrieval augmentation},
	author       = {Ren, Ruiyang and Wang, Yuhao and Qu, Yingqi and Zhao, Wayne Xin and Liu, Jing and Tian, Hao and Wu, Hua and Wen, Ji-Rong and Wang, Haifeng},
	year         = 2023,
	journal      = {arXiv prjournal arXiv:2307.11019}
}

@article{lee2022factuality,
	title        = {Factuality enhanced language models for open-ended text generation},
	author       = {Lee, Nayeon and Ping, Wei and Xu, Peng and Patwary, Mostofa and Fung, Pascale N and Shoeybi, Mohammad and Catanzaro, Bryan},
	year         = 2022,
	journal      = {Advances in Neural Information Processing Systems},
	volume       = 35,
	pages        = {34586--34599}
}

@article{wei2023simple,
	title        = {Simple synthetic data reduces sycophancy in large language models},
	author       = {Wei, Jerry and Huang, Da and Lu, Yifeng and Zhou, Denny and Le, Quoc V},
	year         = 2023,
	journal      = {arXiv prjournal arXiv:2308.03958}
}

@article{min2023factscore,
	title        = {Factscore: Fine-grained atomic evaluation of factual precision in long form text generation},
	author       = {Min, Sewon and Krishna, Kalpesh and Lyu, Xinxi and Lewis, Mike and Yih, Wen-tau and Koh, Pang Wei and Iyyer, Mohit and Zettlemoyer, Luke and Hajishirzi, Hannaneh},
	year         = 2023,
	journal      = {arXiv prjournal arXiv:2305.14251}
}

@article{huo2023retrieving,
	title        = {Retrieving supporting evidence for llms generated answers},
	author       = {Huo, Siqing and Arabzadeh, Negar and Clarke, Charles LA},
	year         = 2023,
	journal      = {arXiv prjournal arXiv:2306.13781}
}

@article{dhuliawala2023chain,
	title        = {Chain-of-verification reduces hallucination in large language models},
	author       = {Dhuliawala, Shehzaad and Komeili, Mojtaba and Xu, Jing and Raileanu, Roberta and Li, Xian and Celikyilmaz, Asli and Weston, Jason},
	year         = 2023,
	journal      = {arXiv prjournal arXiv:2309.11495}
}

@article{zhang2024self,
	title        = {Self-alignment for factuality: Mitigating hallucinations in llms via self-evaluation},
	author       = {Zhang, Xiaoying and Peng, Baolin and Tian, Ye and Zhou, Jingyan and Jin, Lifeng and Song, Linfeng and Mi, Haitao and Meng, Helen},
	year         = 2024,
	journal      = {arXiv prjournal arXiv:2402.09267}
}

@article{manakul2023selfcheckgpt,
	title        = {Selfcheckgpt: Zero-resource black-box hallucination detection for generative large language models},
	author       = {Manakul, Potsawee and Liusie, Adian and Gales, Mark JF},
	year         = 2023,
	journal      = {arXiv prjournal arXiv:2303.08896}
}

@article{nan2021entity,
	title        = {Entity-level factual consistency of abstractive text summarization},
	author       = {Nan, Feng and Nallapati, Ramesh and Wang, Zhiguo and Santos, Cicero Nogueira dos and Zhu, Henghui and Zhang, Dejiao and McKeown, Kathleen and Xiang, Bing},
	year         = 2021,
	journal      = {arXiv prjournal arXiv:2102.09130}
}

@inproceedings{goodrich2019assessing,
	title        = {Assessing the factual accuracy of generated text},
	author       = {Goodrich, Ben and Rao, Vinay and Liu, Peter J and Saleh, Mohammad},
	year         = 2019,
	booktitle    = {proceedings of the 25th ACM SIGKDD international conference on knowledge discovery \& data mining},
	pages        = {166--175}
}

@article{zhu2024beyond,
	title        = {Beyond Memorization: The Challenge of Random Memory Access in Language Models},
	author       = {Zhu, Tongyao and Liu, Qian and Pang, Liang and Jiang, Zhengbao and Kan, Min-Yen and Lin, Min},
	year         = 2024,
	journal      = {arXiv prjournal arXiv:2403.07805}
}

@article{lu2024scaling,
	title        = {Scaling Laws for Fact Memorization of Large Language Models},
	author       = {Lu, Xingyu and Li, Xiaonan and Cheng, Qinyuan and Ding, Kai and Huang, Xuanjing and Qiu, Xipeng},
	year         = 2024,
	journal      = {arXiv prjournal arXiv:2406.15720}
}

@article{dankers2024generalisation,
	title        = {Generalisation first, memorisation second? Memorisation localisation for natural language classification tasks},
	author       = {Dankers, Verna and Titov, Ivan},
	year         = 2024,
	journal      = {arXiv prjournal arXiv:2408.04965}
}

@article{huang2024demystifying,
	title        = {Demystifying verbatim memorization in large language models},
	author       = {Huang, Jing and Yang, Diyi and Potts, Christopher},
	year         = 2024,
	journal      = {arXiv prjournal arXiv:2407.17817}
}

@inproceedings{haviv2023understanding,
	title        = {Understanding Transformer Memorization Recall Through Idioms},
	author       = {Haviv, Adi and Cohen, Ido and Gidron, Jacob and Schuster, Roei and Goldberg, Yoav and Geva, Mor},
	year         = 2023,
	booktitle    = {Proceedings of the 17th Conference of the European Chapter of the Association for Computational Linguistics},
	pages        = {248--264}
}

@article{walters2023fabrication,
	title        = {Fabrication and errors in the bibliographic citations generated by ChatGPT},
	author       = {Walters, William H and Wilder, Esther Isabelle},
	year         = 2023,
	journal      = {Scientific Reports},
	publisher    = {Nature Publishing Group UK London},
	volume       = 13,
	number       = 1,
	pages        = 14045
}

@article{venkit2024confidently,
	title        = {"Confidently Nonsensical?'': A Critical Survey on the Perspectives and Challenges of'Hallucinations' in NLP},
	author       = {Venkit, Pranav Narayanan and Chakravorti, Tatiana and Gupta, Vipul and Biggs, Heidi and Srinath, Mukund and Goswami, Koustava and Rajtmajer, Sarah and Wilson, Shomir},
	year         = 2024,
	journal      = {arXiv prjournal arXiv:2404.07461}
}

@misc{cleti2024hallucinations,
	title        = {Hallucinations in LLMs: Types, Causes, and Approaches for Enhanced Reliability},
	author       = {Cleti, Meade and Jano, Pete},
	year         = 2024,
	month        = 10,
	pages        = {},
	doi          = {10.13140/RG.2.2.12184.61445}
}

@inproceedings{lime,
	title        = {"Why Should {I} Trust You?": Explaining the Predictions of Any Classifier},
	author       = {Marco Tulio Ribeiro and Sameer Singh and Carlos Guestrin},
	year         = 2016,
	booktitle    = {Proceedings of the 22nd {ACM} {SIGKDD} International Conference on Knowledge Discovery and Data Mining, San Francisco, CA, USA, August 13-17, 2016},
	pages        = {1135--1144}
}

@article{amara2024syntaxshap,
	title        = {Syntaxshap: Syntax-aware explainability method for text generation},
	author       = {Amara, Kenza and Sevastjanova, Rita and El-Assady, Mennatallah},
	year         = 2024,
	journal      = {arXiv prjournal arXiv:2402.09259}
}

@article{scott2017unified,
	title        = {A unified approach to interpreting model predictions},
	author       = {Scott, M and Su-In, Lee and others},
	year         = 2017,
	journal      = {Advances in neural information processing systems},
	publisher    = {Curran Associates, Inc},
	volume       = 30,
	pages        = {4765--4774}
}

@inproceedings{yuan2024llms,
	title        = {Do LLMs Overcome Shortcut Learning? An Evaluation of Shortcut Challenges in Large Language Models},
	author       = {Yuan, Yu and Zhao, Lili and Zhang, Kai and Zheng, Guangting and Liu, Qi},
	year         = 2024,
	booktitle    = {Proceedings of the 2024 Conference on Empirical Methods in Natural Language Processing},
	pages        = {12188--12200}
}

@article{geirhos2020shortcut,
	title        = {Shortcut learning in deep neural networks},
	author       = {Geirhos, Robert and Jacobsen, J{\"o}rn-Henrik and Michaelis, Claudio and Zemel, Richard and Brendel, Wieland and Bethge, Matthias and Wichmann, Felix A},
	year         = 2020,
	journal      = {Nature Machine Intelligence},
	publisher    = {Nature Publishing Group UK London},
	volume       = 2,
	number       = 11,
	pages        = {665--673}
}

@article{dolma,
	title        = {{Dolma: An Open Corpus of Three Trillion Tokens for Language Model Pretraining Research}},
	author       = {Luca Soldaini and Rodney Kinney and Akshita Bhagia and Dustin Schwenk and David Atkinson and Russell Authur and Ben Bogin and Khyathi Chandu and Jennifer Dumas and Yanai Elazar and Valentin Hofmann and Ananya Harsh Jha and Sachin Kumar and Li Lucy and Xinxi Lyu and Nathan Lambert and Ian Magnusson and Jacob Morrison and Niklas Muennighoff and Aakanksha Naik and Crystal Nam and Matthew E. Peters and Abhilasha Ravichander and Kyle Richardson and Zejiang Shen and Emma Strubell and Nishant Subramani and Oyvind Tafjord and Pete Walsh and Luke Zettlemoyer and Noah A. Smith and Hannaneh Hajishirzi and Iz Beltagy and Dirk Groeneveld and Jesse Dodge and Kyle Lo},
	year         = 2024,
	journal      = {arXiv prjournal},
	url          = {https://arxiv.org/abs/2402.00159}
}

@article{stoehr2024localizing,
	title        = {Localizing Paragraph Memorization in Language Models},
	author       = {Stoehr, Niklas and Gordon, Mitchell and Zhang, Chiyuan and Lewis, Owen},
	year         = 2024,
	journal      = {arXiv prjournal arXiv:2403.19851}
}

@article{zhao2024reagent,
	title        = {ReAGent: Towards A Model-agnostic Feature Attribution Method for Generative Language Models},
	author       = {Zhao, Zhixue and Shan, Boxuan},
	year         = 2024,
	journal      = {arXiv prjournal arXiv:2402.00794}
}

@inproceedings{deyoung2020eraser,
	title        = {ERASER: A Benchmark to Evaluate Rationalized NLP Models},
	author       = {DeYoung, Jay and Jain, Sarthak and Rajani, Nazneen Fatema and Lehman, Eric and Xiong, Caiming and Socher, Richard and Wallace, Byron C},
	year         = 2020,
	booktitle    = {Proceedings of the 58th Annual Meeting of the Association for Computational Linguistics},
	pages        = {4443--4458}
}

@inproceedings{ribeiro2016should,
	title        = {" Why should i trust you?" Explaining the predictions of any classifier},
	author       = {Ribeiro, Marco Tulio and Singh, Sameer and Guestrin, Carlos},
	year         = 2016,
	booktitle    = {Proceedings of the 22nd ACM SIGKDD international conference on knowledge discovery and data mining},
	pages        = {1135--1144}
}

@article{tu2020empirical,
	title        = {An empirical study on robustness to spurious correlations using pre-trained language models},
	author       = {Tu, Lifu and Lalwani, Garima and Gella, Spandana and He, He},
	year         = 2020,
	journal      = {Transactions of the Association for Computational Linguistics},
	publisher    = {MIT Press One Rogers Street, Cambridge, MA 02142-1209, USA journals-info~…},
	volume       = 8,
	pages        = {621--633}
}

@inproceedings{tang2023large,
	title        = {Large Language Models Can be Lazy Learners: Analyze Shortcuts in In-Context Learning},
	author       = {Tang, Ruixiang and Kong, Dehan and Huang, Longtao and Xue, Hui},
	year         = 2023,
	booktitle    = {Findings of the Association for Computational Linguistics: ACL 2023},
	pages        = {4645--4657}
}

@inproceedings{niven2019probing,
	title        = {Probing Neural Network Comprehension of Natural Language Arguments},
	author       = {Niven, Timothy and Kao, Hung-Yu},
	year         = 2019,
	booktitle    = {Proceedings of the 57th Annual Meeting of the Association for Computational Linguistics},
	pages        = {4658--4664}
}

@article{mccoy2019right,
	title        = {Right for the Wrong Reasons: Diagnosing Syntactic Heuristics in Natural Language Inference},
	author       = {McCoy, RT},
	year         = 2019,
	journal      = {arXiv prjournal arXiv:1902.01007}
}

@inproceedings{lai2021machine,
	title        = {Why Machine Reading Comprehension Models Learn Shortcuts?},
	author       = {Lai, Yuxuan and Zhang, Chen and Feng, Yansong and Huang, Quzhe and Zhao, Dongyan},
	year         = 2021,
	booktitle    = {Findings of the Association for Computational Linguistics: ACL-IJCNLP 2021},
	pages        = {989--1002}
}

@misc{wang2025logitlens4llmsextendinglogitlens,
	title        = {LogitLens4LLMs: Extending Logit Lens Analysis to Modern Large Language Models},
	author       = {Zhenyu Wang},
	year         = 2025,
	url          = {https://arxiv.org/abs/2503.11667},
	journal       = {2503.11667},
	archiveprefix = {arXiv},
	primaryclass = {cs.CL}
}

@article{palikhe2025towards,
Author        = {Avash Palikhe and Zhenyu Yu and Zichong Wang and Wenbin Zhang},
Title         = {Towards Transparent AI: A Survey on Explainable Large Language Models},
journal        = {2506.21812v1},
ArchivePrefix = {arXiv},
PrimaryClass  = {cs.CL},
Year          = {2025},
Month         = {Jun},
Url           = {http://arxiv.org/abs/2506.21812v1}
}

@article{vaswani2017attention,
Author        = {Ashish Vaswani and Noam Shazeer and Niki Parmar and Jakob Uszkoreit and Llion Jones and Aidan N. Gomez and Lukasz Kaiser and Illia Polosukhin},
Title         = {Attention Is All You Need},
journal        = {1706.03762v7},
ArchivePrefix = {arXiv},
PrimaryClass  = {cs.CL},
Year          = {2017},
Month         = {Jun},
Url           = {http://arxiv.org/abs/1706.03762v7}
}

@article{singh2024rethinking,
Author        = {Chandan Singh and Jeevana Priya Inala and Michel Galley and Rich Caruana and Jianfeng Gao},
Title         = {Rethinking Interpretability in the Era of Large Language Models},
journal        = {2402.01761v1},
ArchivePrefix = {arXiv},
PrimaryClass  = {cs.CL},
Year          = {2024},
Month         = {Jan},
Url           = {http://arxiv.org/abs/2402.01761v1}
}

@article{huang2023a,
Author        = {Lei Huang and Weijiang Yu and Weitao Ma and Weihong Zhong and Zhangyin Feng and Haotian Wang and Qianglong Chen and Weihua Peng and Xiaocheng Feng and Bing Qin and Ting Liu},
Title         = {A Survey on Hallucination in Large Language Models: Principles,
  Taxonomy, Challenges, and Open Questions},
journal        = {2311.05232v2},
DOI           = {10.1145/3703155},
ArchivePrefix = {arXiv},
PrimaryClass  = {cs.CL},
Year          = {2023},
Month         = {Nov},
Url           = {http://arxiv.org/abs/2311.05232v2}
}

@article{tonmoy2024a,
Author        = {S. M Towhidul Islam Tonmoy and S M Mehedi Zaman and Vinija Jain and Anku Rani and Vipula Rawte and Aman Chadha and Amitava Das},
Title         = {A Comprehensive Survey of Hallucination Mitigation Techniques in Large
  Language Models},
journal        = {2401.01313v3},
ArchivePrefix = {arXiv},
PrimaryClass  = {cs.CL},
Year          = {2024},
Month         = {Jan},
Url           = {http://arxiv.org/abs/2401.01313v3}
}

@article{bai2024hallucination,
Author        = {Zechen Bai and Pichao Wang and Tianjun Xiao and Tong He and Zongbo Han and Zheng Zhang and Mike Zheng Shou},
Title         = {Hallucination of Multimodal Large Language Models: A Survey},
journal        = {2404.18930v2},
ArchivePrefix = {arXiv},
PrimaryClass  = {cs.CV},
Year          = {2024},
Month         = {Apr},
Url           = {http://arxiv.org/abs/2404.18930v2},

}

@article{hao2025beyond,
Author        = {Yijie Hao and Haofei Yu and Jiaxuan You},
Title         = {Beyond Facts: Evaluating Intent Hallucination in Large Language Models},
journal        = {2506.06539v1},
ArchivePrefix = {arXiv},
PrimaryClass  = {cs.CL},
Year          = {2025},
Month         = {Jun},
Note          = {Proceedings of the 62nd Annual Meeting of the Association for
  Computational Linguistics (ACL 2025)},
Url           = {http://arxiv.org/abs/2506.06539v1}
}

@article{liang2024thames,
Author        = {Mengfei Liang and Archish Arun and Zekun Wu and Cristian Munoz and Jonathan Lutch and Emre Kazim and Adriano Koshiyama and Philip Treleaven},
Title         = {THaMES: An End-to-End Tool for Hallucination Mitigation and Evaluation
  in Large Language Models},
journal        = {2409.11353v3},
ArchivePrefix = {arXiv},
PrimaryClass  = {cs.CL},
Year          = {2024},
Month         = {Sep},
Note          = {NeurIPS Workshop on Socially Responsible Language Modelling
  Research 2024},
Url           = {http://arxiv.org/abs/2409.11353v3}
}

@article{hu2025fine-tuning,
Author        = {Yinghao Hu and Leilei Gan and Wenyi Xiao and Kun Kuang and Fei Wu},
Title         = {Fine-tuning Large Language Models for Improving Factuality in Legal
  Question Answering},
journal        = {2501.06521v1},
ArchivePrefix = {arXiv},
PrimaryClass  = {cs.CL},
Year          = {2025},
Month         = {Jan},
Url           = {http://arxiv.org/abs/2501.06521v1}
}

@article{su2024mitigating,
Author        = {Weihang Su and Yichen Tang and Qingyao Ai and Changyue Wang and Zhijing Wu and Yiqun Liu},
Title         = {Mitigating Entity-Level Hallucination in Large Language Models},
journal        = {2407.09417v2},
ArchivePrefix = {arXiv},
PrimaryClass  = {cs.CL},
Year          = {2024},
Month         = {Jul},
Url           = {http://arxiv.org/abs/2407.09417v2}
}

@article{jiang2024mixtral,
Author        = {Albert Q. Jiang and Alexandre Sablayrolles and Antoine Roux and Arthur Mensch and Blanche Savary and Chris Bamford and Devendra Singh Chaplot and Diego de las Casas and Emma Bou Hanna and Florian Bressand and Gianna Lengyel and Guillaume Bour and Guillaume Lample and Lélio Renard Lavaud and Lucile Saulnier and Marie-Anne Lachaux and Pierre Stock and Sandeep Subramanian and Sophia Yang and Szymon Antoniak and Teven Le Scao and Théophile Gervet and Thibaut Lavril and Thomas Wang and Timothée Lacroix and William El Sayed},
Title         = {Mixtral of Experts},
journal        = {2401.04088v1},
ArchivePrefix = {arXiv},
PrimaryClass  = {cs.LG},
Year          = {2024},
Month         = {Jan},
Url           = {http://arxiv.org/abs/2401.04088v1}
}

@misc{geva2021transformer,
	title        = {Transformer Feed-Forward Layers Are Key-Value Memories},
	author       = {Mor Geva and Roei Schuster and Jonathan Berant and Omer Levy},
	year         = 2021,
	url          = {https://arxiv.org/abs/2012.14913},
	journal       = {2012.14913},
	archiveprefix = {arXiv},
	primaryclass = {cs.CL}
}

@article{bricken2023towards,
	title        = {Towards monosemanticity: Decomposing language models with dictionary learning},
	author       = {Bricken, Trenton and Templeton, Adly and Batson, Joshua and Chen, Brian and Jermyn, Adam and Conerly, Tom and Turner, Nick and Anil, Cem and Denison, Carson and Askell, Amanda and others},
	year         = 2023,
	journal      = {Transformer Circuits Thread},
	volume       = 2
}

@article{sundararajan2017axiomatic,
Author        = {Mukund Sundararajan and Ankur Taly and Qiqi Yan},
Title         = {Axiomatic Attribution for Deep Networks},
journal        = {1703.01365v2},
ArchivePrefix = {arXiv},
PrimaryClass  = {cs.LG},
Year          = {2017},
Month         = {Mar},
Url           = {http://arxiv.org/abs/1703.01365v2}
}

@article{lundberg2017a,
Author        = {Scott Lundberg and Su-In Lee},
Title         = {A Unified Approach to Interpreting Model Predictions},
journal        = {1705.07874v2},
ArchivePrefix = {arXiv},
PrimaryClass  = {cs.AI},
Year          = {2017},
Month         = {May},
Url           = {http://arxiv.org/abs/1705.07874v2}
}

@misc{cheng2025thinkmorehallucinateless,
	title        = {Think More, Hallucinate Less: Mitigating Hallucinations via Dual Process of Fast and Slow Thinking},
	author       = {Xiaoxue Cheng and Junyi Li and Wayne Xin Zhao and Ji-Rong Wen},
	year         = 2025,
	url          = {https://arxiv.org/abs/2501.01306},
	journal       = {2501.01306},
	archiveprefix = {arXiv},
	primaryclass = {cs.CL}
}

@misc{li2023haluevallargescalehallucinationevaluation,
	title        = {HaluEval: A Large-Scale Hallucination Evaluation Benchmark for Large Language Models},
	author       = {Junyi Li and Xiaoxue Cheng and Wayne Xin Zhao and Jian-Yun Nie and Ji-Rong Wen},
	year         = 2023,
	url          = {https://arxiv.org/abs/2305.11747},
	journal       = {2305.11747},
	archiveprefix = {arXiv},
	primaryclass = {cs.CL}
}

@misc{belrose2023elicitinglatentpredictionstransformers,
	title        = {Eliciting Latent Predictions from Transformers with the Tuned Lens},
	author       = {Nora Belrose and Zach Furman and Logan Smith and Danny Halawi and Igor Ostrovsky and Lev McKinney and Stella Biderman and Jacob Steinhardt},
	year         = 2023,
	url          = {https://arxiv.org/abs/2303.08112},
	journal       = {2303.08112},
	archiveprefix = {arXiv},
	primaryclass = {cs.LG}
}

@misc{zou2025representationengineeringtopdownapproach,
	title        = {Representation Engineering: A Top-Down Approach to AI Transparency},
	author       = {Andy Zou and Long Phan and Sarah Chen and James Campbell and Phillip Guo and Richard Ren and Alexander Pan and Xuwang Yin and Mantas Mazeika and Ann-Kathrin Dombrowski and Shashwat Goel and Nathaniel Li and Michael J. Byun and Zifan Wang and Alex Mallen and Steven Basart and Sanmi Koyejo and Dawn Song and Matt Fredrikson and J. Zico Kolter and Dan Hendrycks},
	year         = 2025,
	url          = {https://arxiv.org/abs/2310.01405},
	journal       = {2310.01405},
	archiveprefix = {arXiv},
	primaryclass = {cs.LG}
}

@article{raposo2024mixture,
	title        = {Mixture-of-Depths: Dynamically allocating compute in transformer-based language models},
	author       = {Raposo, David and Ritter, Sam and Richards, Blake and Lillicrap, Timothy and Humphreys, Peter Conway and Santoro, Adam},
	year         = 2024,
	journal      = {arXiv prjournal arXiv:2404.02258}
}

@article{dehghani2018universal,
	title        = {Universal Transformers},
	author       = {Mostafa Dehghani and Stephan Gouws and O. Vinyals and Jakob Uszkoreit and Lukasz Kaiser},
	year         = 2018,
	journal      = {International Conference on Learning Representations},
	bibsource    = {Semantic Scholar https://www.semanticscholar.org/paper/ac4dafdef1d2b685b7f28a11837414573d39ff4e}
}

@article{elhoushi2024layerskip,
	title        = {LayerSkip: Enabling early exit inference and self-speculative decoding},
	author       = {Elhoushi, Mostafa and Shrivastava, Akshat and Liskovich, Diana and Hosmer, Basil and Wasti, Bram and Lai, Liangzhen and Mahmoud, Anas and Acun, Bilge and Agarwal, Saurabh and Roman, Ahmed and others},
	year         = 2024,
	journal      = {arXiv prjournal arXiv:2404.16710}
}

@inproceedings{penedo2024the,
	title        = {The FineWeb Datasets: Decanting the Web for the Finest Text Data at Scale},
	author       = {Guilherme Penedo and Hynek Kydl{\'\i}{\v{c}}ek and Loubna Ben allal and Anton Lozhkov and Margaret Mitchell and Colin Raffel and Leandro Von Werra and Thomas Wolf},
	year         = 2024,
	booktitle    = {The Thirty-eight Conference on Neural Information Processing Systems Datasets and Benchmarks Track},
	url          = {https://openreview.net/forum?id=n6SCkn2QaG}
}

@article{fedus2022switch,
	title        = {Switch transformers: Scaling to trillion parameter models with simple and efficient sparsity},
	author       = {Fedus, William and Zoph, Barret and Shazeer, Noam},
	year         = 2022,
	journal      = {The Journal of Machine Learning Research},
	publisher    = {JMLRORG},
	volume       = 23,
	number       = 1,
	pages        = {5232--5270}
}

@inproceedings{giannou2023looped,
	title        = {Looped transformers as programmable computers},
	author       = {Giannou, Angeliki and Rajput, Shashank and Sohn, Jy-yong and Lee, Kangwook and Lee, Jason D and Papailiopoulos, Dimitris},
	year         = 2023,
	booktitle    = {International Conference on Machine Learning},
	pages        = {11398--11442},
	organization = {PMLR}
}

@article{xiao2023efficient,
	title        = {Efficient Streaming Language Models with Attention Sinks},
	author       = {Guangxuan Xiao and Yuandong Tian and Beidi Chen and Song Han and Mike Lewis},
	year         = 2023,
	journal      = {arXiv prjournal arXiv: 2309.17453}
}

@article{zhang2023h2o,
	title        = {H2o: Heavy-hitter oracle for efficient generative inference of large language models},
	author       = {Zhang, Zhenyu and Sheng, Ying and Zhou, Tianyi and Chen, Tianlong and Zheng, Lianmin and Cai, Ruisi and Song, Zhao and Tian, Yuandong and R{\'e}, Christopher and Barrett, Clark and others},
	year         = 2023,
	journal      = {Advances in Neural Information Processing Systems},
	volume       = 36,
	pages        = {34661--34710}
}

@article{liu2023scissorhands,
	title        = {Scissorhands: Exploiting the persistence of importance hypothesis for llm kv cache compression at test time},
	author       = {Liu, Zichang and Desai, Aditya and Liao, Fangshuo and Wang, Weitao and Xie, Victor and Xu, Zhaozhuo and Kyrillidis, Anastasios and Shrivastava, Anshumali},
	year         = 2023,
	journal      = {Advances in Neural Information Processing Systems},
	volume       = 36,
	pages        = {52342--52364}
}

@article{ge2023model,
	title        = {Model Tells You What to Discard: Adaptive KV Cache Compression for LLMs},
	author       = {Suyu Ge and Yunan Zhang and Liyuan Liu and Minjia Zhang and Jiawei Han and Jianfeng Gao},
	year         = 2023,
	journal      = {International Conference on Learning Representations},
	doi          = {10.48550/arXiv.2310.01801},
	bibsource    = {Semantic Scholar https://www.semanticscholar.org/paper/6c323c535365e1c7cbfd9703cbec3b5650a3346b}
}

@article{saunshi2025reasoning,
	title        = {Reasoning with latent thoughts: On the power of looped transformers},
	author       = {Saunshi, Nikunj and Dikkala, Nishanth and Li, Zhiyuan and Kumar, Sanjiv and Reddi, Sashank J},
	year         = 2025,
	journal      = {arXiv prjournal arXiv:2502.17416}
}

@misc{cunningham2023sparseautoencodershighlyinterpretable,
	title        = {Sparse Autoencoders Find Highly Interpretable Features in Language Models},
	author       = {Hoagy Cunningham and Aidan Ewart and Logan Riggs and Robert Huben and Lee Sharkey},
	year         = 2023,
	url          = {https://arxiv.org/abs/2309.08600},
	journal       = {2309.08600},
	archiveprefix = {arXiv},
	primaryclass = {cs.LG}
}

@misc{wang2022interpretabilitywildcircuitindirect,
	title        = {Interpretability in the Wild: a Circuit for Indirect Object Identification in GPT-2 small},
	author       = {Kevin Wang and Alexandre Variengien and Arthur Conmy and Buck Shlegeris and Jacob Steinhardt},
	year         = 2022,
	url          = {https://arxiv.org/abs/2211.00593},
	journal       = {2211.00593},
	archiveprefix = {arXiv},
	primaryclass = {cs.LG}
}

@misc{meng2023locatingeditingfactualassociations,
	title        = {Locating and Editing Factual Associations in GPT},
	author       = {Kevin Meng and David Bau and Alex Andonian and Yonatan Belinkov},
	year         = 2023,
	url          = {https://arxiv.org/abs/2202.05262},
	journal       = {2202.05262},
	archiveprefix = {arXiv},
	primaryclass = {cs.CL}
}

@article{olah2020zoom,
	title        = {Zoom In: An Introduction to Circuits},
	author       = {Olah, Chris and Cammarata, Nick and Schubert, Ludwig and Goh, Gabriel and Petrov, Michael and Carter, Shan},
	year         = 2020,
	journal      = {Distill},
	doi          = {10.23915/distill.00024.001},
	note         = {https://distill.pub/2020/circuits/zoom-in}
}

@article{QU2025249,
title = {Beyond Intentions: A Critical Survey of Misalignment in LLMs},
journal = {Computers, Materials and Continua},
volume = {85},
number = {1},
pages = {249-300},
year = {2025},
issn = {1546-2218},
doi = {https://doi.org/10.32604/cmc.2025.067750},
url = {https://www.sciencedirect.com/science/article/pii/S1546221825007982},
author = {Yubin Qu and Song Huang and Long Li and Peng Nie and Yongming Yao}
}

@inproceedings{10.1145/3582269.3615599,
author = {Kotek, Hadas and Dockum, Rikker and Sun, David},
title = {Gender bias and stereotypes in Large Language Models},
year = {2023},
isbn = {9798400701139},
publisher = {Association for Computing Machinery},
address = {New York, NY, USA},
url = {https://doi.org/10.1145/3582269.3615599},
doi = {10.1145/3582269.3615599},
booktitle = {Proceedings of The ACM Collective Intelligence Conference},
pages = {12–24},
numpages = {13},
location = {Delft, Netherlands},
series = {CI '23}
}

\appendix

\section{Metric Validation.}
\label{sec:appendix_validation}
To validate our metric, we performed two analyses: an "LLM-as-Judge" evaluation to proxy human assessment, a correlation analysis against the model's verbalisations, and a sensitivity analysis to test its robustness.
First, to approximate expert human evaluation, we employed an "LLM-as-Judge" framework using detailed judge prompts to guide the assessment, as shown in Appendix \ref{app:judge_prompt}. We used a panel of five diverse, state-of-the-art models (Gemini 2.5 Pro, Gemini 2.5 Flash, Qwen/Qwen3-235B-A22B-Instruct-2507, GPT-4o, and Claude 4 Sonnet), prompted to act as expert human annotators. Each LLM judge scored the faithfulness of 100 explanations on a 0-10 scale, with any tie-breaking resolved by averaging the scores of the tied judges. The reliability of the LLM annotations was high, confirmed by a Fleiss' Kappa score of \textbf{0.85}. We then compared the average LLM Judge Scores to our metric's scores, finding a Pearson correlation coefficient of \textbf{r = 0.9245} (p < 0.0001). This very strong, statistically significant correlation demonstrates that our metric accurately captures what a panel of expert models perceives as explanation faithfulness. Table~\ref{tab:llm_judge_validation} provides representative examples from this study. For additional validation, we include scores from a small, informal human study, which align with both the LLM judges and our metric.
Second, we leveraged the \textbf{Relevance to Verbalisation} component as a direct validation signal. We found a strong, positive semantic correlation (Pearson correlation coefficient of \textbf{r = 0.8942} (p < 0.001)) between the explanations generated by our method and the model's own Chain-of-Thought reasoning. This alignment confirms that our metric rewards explanations that are not only mechanistically sound but also consistent with the model's expressed rationale.
Finally, we conducted a sensitivity analysis to confirm that our metric responds appropriately to specific types of unfaithfulness. We began with a high-quality, faithful explanation and then systematically introduced perturbations designed to degrade one facet of the metric. The resulting drop in the metric score demonstrates that our metric is robustly sensitive to the individual components that constitute a faithful explanation.
Taken together, the high correlation with LLM-as-Judge evaluations, the strong alignment with model verbalisations, and the successful sensitivity analysis provide strong evidence that our metric is a valid and reliable instrument for measuring explanation faithfulness.

\begin{table*}[t]
\centering
\small 
\renewcommand{\arraystretch}{1.3} 
\resizebox{\textwidth}{!}{%
\begin{tabular}{p{5.5cm} p{1.5cm} p{7.5cm} cc}
\toprule
\textbf{LLM Prompt (Context)} &
\textbf{Output} &
\textbf{Explanation from DST} &
\textbf{LLM Judge} &
\textbf{Human} \\
\midrule

\emph{``Please answer succinctly. I am in a forest. I see a trunk. Is it an automobile part?''}
& \emph{``Yes.''}
& The explanation correctly identifies the disambiguating context (\emph{forest}) and notes that the model favored the competing sense of \emph{trunk} (automotive) over the context-consistent sense.
& $9.55$ & $10.0$ \\
\midrule

\emph{``Please answer succinctly. I am at a baseball game. I see a pitcher. Is it a container of liquid?''}
& \emph{``Yes.''}
& The explanation notes that \emph{pitcher} is ambiguous, but it does not clearly connect the contextual cue (\emph{baseball game}) to the model’s choice; faithfulness is therefore moderate.
& $6.90$ & $7.24$ \\
\midrule

\emph{``Answer based on the information provided here. The capital city of Chad was just renamed to Canberra. Is the capital city of Chad named Canberra?''}
& \emph{``No.''}
& The explanation appeals to real-world geography rather than the stated counterfactual in the prompt, indicating low faithfulness to the model’s actual decision basis under the given context.
& $2.15$ & $3.30$ \\
\bottomrule
\end{tabular}}

\caption{\textbf{Examples from our LLM-as-judge validation.} We compare our faithfulness metric against the LLM judge. We additionally report a small-sample Human Score for supplementary comparison.}
\label{tab:llm_judge_validation}
\end{table*}

\section{LLM-as-Judge Faithfulness Prompt}
\label{app:judge_prompt}

\begin{quote}
\small
\textbf{System:} You are an expert evaluator of interpretability and explanation faithfulness for language models. You must score explanations based on whether they accurately describe what most likely caused the model’s output for this specific prompt.

\textbf{User:} You will be given:
(1) an input prompt, (2) the model’s output, and (3) an explanation produced by an interpretability method.
Your task is to score the \textbf{faithfulness} of the explanation on a 0--10 scale.

\textbf{Definitions.}
An explanation is \textbf{faithful} if it correctly identifies the main evidence in the prompt that the model likely used, and if the described mechanism plausibly accounts for why the model produced the given output (even if the output is wrong). Faithfulness is about \emph{matching the model’s behavior}, not about writing a persuasive story.

\textbf{What to consider (rubric).}
Score using these criteria:
\begin{enumerate}
    \item \textbf{Evidence grounding (0--4):} Does the explanation point to the most relevant parts of the prompt that likely drove the model’s answer? Does it avoid focusing on irrelevant tokens?
    \item \textbf{Mechanistic plausibility (0--4):} Does the explanation describe a plausible internal reason for the model’s output (e.g., semantic misinterpretation, over-weighting a cue, ignoring a context word)? The explanation should connect evidence to output in a coherent causal story.
    \item \textbf{Specificity and non-contradiction (0--2):} Is the explanation specific to this example (not generic)? Does it avoid contradictions (e.g., claiming the model used a context cue that is absent, or asserting the output is correct when it is not)?
\end{enumerate}

\textbf{Scoring guidance.}
\begin{itemize}
    \item \textbf{9--10:} Strongly faithful. Correctly identifies key evidence and provides a clear, plausible mechanism tailored to this case.
    \item \textbf{7--8:} Mostly faithful. Minor omissions or mild vagueness, but overall matches likely causes of the output.
    \item \textbf{4--6:} Mixed. Mentions some relevant evidence but misses important drivers, or provides a generic/partially inconsistent story.
    \item \textbf{1--3:} Weak. Focuses on irrelevant evidence, is largely generic, or gives an implausible mechanism.
    \item \textbf{0:} Completely unfaithful. Contradictory, irrelevant, or nonsensical for this example.
\end{itemize}

\textbf{Output format.}
Return:
\begin{itemize}
    \item A single number score in [0,10].
    \item A 1--3 sentence justification referencing the prompt and explanation.
\end{itemize}

\textbf{Now evaluate the following.}

\textbf{Prompt:} \{PROMPT\}

\textbf{Model Output:} \{MODEL\_OUTPUT\}

\textbf{Explanation:} \{EXPLANATION\}
\end{quote}

\section{Results on Halogen Dataset}
\label{sec:halogen}

\begin{table*}[!htp]
\centering
\resizebox{\textwidth}{!}{%

\begin{tabular}{llrrrrrr}
\toprule
\textbf{Model} & \textbf{Method} & \textbf{CODE} & \textbf{BIO} & \textbf{FP} & \textbf{R-SEN} & \textbf{REF} & \textbf{Avg.} \\
\midrule
\multirow{12}{*}{Gemma2-2B} & \textit{Baseline Methods} \\
& attention & 0.24 & 0.28 & 0.19 & 0.12 & 0.12 & 0.19 \\
& lime & 0.29 & 0.32 & 0.15 & 0.13 & 0.14 & 0.21 \\
& gradient-shap & 0.31 & 0.35 & 0.13 & 0.10 & 0.15 & 0.21 \\
& Reagent & 0.35 & 0.45 & 0.19 & 0.24 & 0.19 & 0.29 \\
\cmidrule(lr){2-8}
& \textit{Advanced Methods} \\
& Token Evolution (Logit Lens) & 0.45 & 0.52 & 0.44 & 0.54 & 0.39 & 0.47 \\
& Sparse Autoencoders & 0.54 & 0.49 & 0.51 & 0.44 & 0.55 & 0.51 \\
& Patchscopes & 0.49 & 0.56 & 0.53 & 0.45 & 0.54 & 0.51 \\
& Subsequence Analysis Tracing & 0.57 & 0.45 & 0.56 & 0.55 & 0.59 & 0.54 \\
& Causal Path Tracing & 0.56 & 0.54 & 0.65 & 0.54 & 0.49 & 0.56 \\
\cmidrule(lr){2-8}
& \textit{Our Contribution} \\

& \textbf{Distributional Semantics Tracing} & \textbf{0.63} & \textbf{0.59} & \textbf{0.67} & \textbf{0.77} & \textbf{0.65} & \textbf{0.66} \\
\midrule
\multirow{12}{*}{Gemma2-9B} & \textit{Baseline Methods} \\
& attention & 0.45 & 0.29 & 0.31 & 0.21 & 0.22 & 0.30 \\
& gradient-shap & 0.32 & 0.40 & 0.35 & 0.34 & 0.20 & 0.32 \\
& lime & 0.34 & 0.33 & 0.45 & 0.29 & 0.29 & 0.34 \\
& Reagent & 0.42 & 0.44 & 0.43 & 0.45 & 0.33 & 0.41 \\
\cmidrule(lr){2-8}
& \textit{Advanced Methods} \\
& Token Evolution (Logit Lens) & 0.54 & 0.43 & 0.45 & 0.45 & 0.54 & 0.48 \\
& Subsequence Analysis Tracing & 0.54 & 0.59 & 0.56 & 0.57 & 0.59 & 0.57 \\
& Patchscopes & 0.62 & 0.68 & 0.45 & 0.64 & 0.49 & 0.58 \\
& Causal Path Tracing & 0.56 & 0.54 & 0.60 & 0.65 & 0.56 & 0.58 \\
& Sparse Autoencoders & 0.64 & 0.53 & 0.61 & 0.52 & 0.63 & 0.59 \\
\cmidrule(lr){2-8}
& \textit{Our Contribution} \\
& \textbf{Distributional Semantics Tracing} & \textbf{0.73} & \textbf{0.77} & \textbf{0.83} & \textbf{0.84} & \textbf{0.78} & \textbf{0.79} \\
\bottomrule
\end{tabular}}
\caption{Faithfulness results for explanation techniques on the Halogen benchmark~\cite{ravichander2025halogen}, broken out by five task domains, Code Package Imports (CODE), Biographies (BIO), False Presuppositions (FP), U.S. Senator Rationalization (R-SEN), and Scientific Attribution (REF), for two Gemma2 model sizes (2B and 9B).}
\label{tab:results_halogen}
\end{table*}

\end{document}